\documentclass[a4paper,fleqn]{cas-sc} 

\usepackage[numbers]{natbib} 

\usepackage{amsmath}
\usepackage{amssymb}
\usepackage{amsfonts}
\usepackage{amsthm}
\usepackage[utf8]{inputenc}
\usepackage[ruled, linesnumbered]{algorithm2e}
\usepackage{placeins}
\usepackage[raggedrightboxes]{ragged2e}

\usepackage[T1]{fontenc}

\makeatletter
\newcommand*{\rom}[1]{\expandafter\@slowromancap\romannumeral #1@}
\makeatother

\newtheoremstyle{break}%
{}{}%
{}{}%
{\bfseries}{.}
{\newline}{}

\theoremstyle{break}
\newtheorem{exmp}{Example}

\def\tsc#1{\csdef{#1}{\textsc{\lowercase{#1}}\xspace}}
\tsc{WGM}
\tsc{QE}
\tsc{EP}
\tsc{PMS}
\tsc{BEC}
\tsc{DE}

\begin{document}
\let\WriteBookmarks\relax
\def\floatpagepagefraction{1}
\def\textpagefraction{.001}

\renewcommand{\aboverulesep}{.2pt}
\renewcommand{\belowrulesep}{1pt}

\shorttitle{Application of Context-dependent Biosignals Recognition to Control a  Hand Prosthesis}

\shortauthors{P. Trajdos, M. Kurzynski} 

\title [mode = title]{Application of Context-dependent Interpretation of Biosignals Recognition to Control a Bionic Multifunctional Hand Prosthesis}
\tnotemark[1,2]

%

%
\author[1]{Pawel Trajdos}[type=editor,
                        auid=000,bioid=1,
                        prefix=,
                        orcid=0000-0002-4337-6847]

\cormark[1]


\ead{pawel.trajdos@pwr.edu.pl}


\credit{Conceptualization (50\%),Data curation, Formal analysis, Investigation, Methodology, Project Administration, Software, Validation, Visualization, Writing -- original draft (70\%), Writing -- review \& editing (70\%)}

\affiliation[1]{organization={Wroclaw University of Science and Technology},
    addressline={Wybrzeze Wyspianskiego 27}, 
    city={Wroclaw},
    postcode={50-370}, 
    country={Poland}}

\author[1]{Marek Kurzynski}[
orcid=0000-0002-0401-2725,
bioid=2
]

\credit{Conceptualization (50\%), Supervision, Writing -- original draft (30\%), Writing -- review \& editing (30\%) } 
\ead{marek.kurzynski@pwr.edu.pl}

\cortext[cor1]{Corresponding author}

%

\begin{abstract}
The paper presents an original method for controlling a surface-electromyography-driven (sEMG) prosthesis. A context-dependent recognition system is proposed in which the same class of sEMG signals may have a different interpretation, depending on the context. This allowed the repertoire of performed movements to be increased. The proposed structure of the context-dependent recognition system includes unambiguously defined decision sequences covering the overall action of the prosthesis, i.e. the so-called boxes.
Because the boxes are mutually isolated environments, each box has its own interpretation of the recognition result, as well as a separate local-recognition-task-focused classifier.

Due to the freedom to assign contextual meanings to classes of biosignals, the construction procedure of the classifier can be optimised in terms of the local classification quality in a given box or the classification quality of the entire system. In the paper, two optimisation problems are formulated, differing in the adopted constraints on optimisation variables, with the methods of solving the problems based on an exhaustive search and an evolutionary algorithm, being developed. 

Experimental studies were conducted using signals from 1 able-bodied person with simulation of amputation and 10 volunteers with transradial amputations. The study compared the classical recognition system and the context-dependent system for various classifier models. An unusual testing strategy was adopted in the research, taking into account the specificity of the considered recognition task, with two original quality measures resulting from this scheme then being applied. The results obtained confirm the hypothesis that the application of the context-dependent classifier led to an improvement in classification quality.  
\end{abstract}


\maketitle
\printcredits

\subsubsection*{Acknowledgments.}
 This work is supported by the National Center for Research and Development (\href{www.ncbr.gov.pl}{www.ncbr.gov.pl}) through project no. \textbf{ /0018/2020-00} within\textit{''Things are for people''} program. The authors have no conflict of interest to declare.

The authors thank Dr. Andrzej Wolczowski for valuable discussions, Dr. Jerzy Witkowski for the design and manufacture of the sEMG sensors and Dr. Michal Bledowski for the application for the recording of the sEMG signals \cite{Wolczowski2017}.

\clearpage

\section{Introduction}\label{sec:itroduction}

The hand is a universal organ that allows the gripping and carrying of various objects and, thanks to receptors, is the source of complex tactile sensations. The dexterity and versatility of the hand as a manipulator are excellent. We can use our hands not only to make various gestures, securely grip objects of any shape, and precisely manipulate very small things but also, for example, to play musical instruments or handle advanced technical devices.
The loss of a hand dramatically worsens the quality of a human's life, as simple activities, such as fastening a button or tying a shoelace, become a problem. The loss of both hands makes independent functioning practically impossible and condemns the person to constant care.
Restoring even a substitute hand to such people makes their lives less burdensome. Hand transplants face significant limitations due to the complicated and expensive transplantation procedure, difficulties in finding a suitable donor, and the need to use immunosuppression for the rest of the subject's life~\cite{Kay2023}. 
Although hand transplantology is making great progress, it is still not a widely used medical procedure~\cite{Milek2023}. An alternative is to equip the subject with the so-called bionic hand prosthesis, i.e. a prosthesis controlled by biological signals~\cite{Ghadage2023}.

Although the first mentions of the use of prostheses come from antiquity, active body-powered prosthetic hands first appeared in the 19\textsuperscript{th} century. Over the years, they have evolved into active externally powered prostheses. In the 1950s, the concept of controlling an active prosthetic hand using an electromyographic (EMG) signal was introduced \cite{Piazza2019}. This opened the way for the intensive development of advanced myoelectrically controlled anthropomorphic upper limb prostheses, which is still ongoing~\cite{Chen2023}.

The EMG signal is generated in the skeletal muscles during contractions that represent neuromuscular activities.  A common method of obtaining the EMG signal is surface electromyography (sEMG),  in which the biosignal is recorded from the surface of the body. Surface EMG is a non-invasive biosignal acquisition technique, unlike intramuscular EMG (iEMG), where electrodes are inserted through the skin into muscle tissue. In the case of a transradial amputation or congenital hand defects, the sEMG signal can be obtained from the surface of the forearm stump. This sEMG signal, because there are many muscles in the forearm that cause movements of the fingers and wrist, is successfully used to control the bioprosthesis. Residual muscles in the stump are still under the control of the subject's nervous system, so despite the lack of a hand, the intent to move causes their activity \cite{Yadav2023}.

Nowadays, in the myoelectric control of the upper limb prosthesis, two basic approaches are distinguished \cite{Parajuli2019}:

\begin{enumerate}
\item Non-recognition-based control. In the simplest conventional non-recognition method implementing the on-off scheme, which is commonly used in commercial (manufactured) myoelectric bioprostheses, two active sensors are located above a pair of residual antagonist muscles (e.g. wrist flexor/extensor in the forearm). The extension / flexion of the phantom wrist can, for example, be mapped into an opening/closing grip, respectively, which means the ability to control one degree of freedom (DoF) of the prosthesis \cite{Hahne2020}.  
This simple on-off control scheme can be easily extended to control multiple DoFs by introducing an additional user-generated signal (e.g. co-contraction) that switches control between the DoFs. Switching can be manual (a special button, e.g. Bebionic hand) or a dedicated mobile application (e.g. iLimb hand)), which is a common semi-automatic solution in commercial prostheses. Measurement of the average sEMG signal's amplitude and, depending on the value, control of the selected physical quantity characterising the movement of the prosthesis (e.g. force, velocity, position, or any function thereof) lead to a proportional control scheme \cite{Chen2023}.

\item Recognition-based control. Myoelectric pattern recognition to decode the intent of the user is the most advanced approach to controlling a powered bioprosthesis. 
In this scheme, the desired classes (from a discrete and finite set) of
movements (grips or manipulations) are discriminated on the basis of sEMG signal patterns by the recognition system, and the variety of prosthesis functions depends directly on the classification performance \cite{Campbell2020}.

\end{enumerate}  

Since the topic of this paper falls under the second approach, we will devote a little more attention to recognition-based control methods.

In the pattern recognition scheme for the control of upper limb prosthesis, it is first the source (user) that generates an object (user's intent of movement), which has a formal representation (EMG signal) subject to observation (sEMG signal), and then the recognition system decides on the class of the object (type of prosthesis movement) based on this observation.
The effectiveness of the prosthesis control process and thus the usefulness of the prosthesis in supporting the everyday life of the user, depend on the quality of the above decision-making scheme.

Ensuring a high quality of the recognition process is a difficult challenge, because this quality depends on many factors that are not always under our control. Therefore, it is reasonable to assume that the myoelectric control of the upper limb prosthesis at the decision level is accompanied by uncertainty conditions, which means that the possibility of errors is inherently inscribed in the scheme of operation of the recognition system. However, designers of bioprosthesis control systems must do everything possible to keep the number (or more generally the probability) of these errors as low as possible. Therefore, the research community aims to develop a recognition-based bioprosthesis control methodology that is inspired by this imperative.

Due to this, for the past two decades there have been numerous papers presenting the application of various methods of classification, feature selection/reduction, and feature extraction (see, for example \cite{Chen2023}), each of which brings added value to this application area.
These papers refer to the so-called academic trend in the development of modern upper limb bioprostheses, because EMG pattern recognition schemes have only recently been deployed in commercial prostheses (the Coapt system and Ottobock’s Myo Plus) \cite{Mendez2021}.

The paper presents an original control system for a hand bioprosthesis based on the pattern recognition paradigm. 
The method proposed in the paper concerns the new decision scheme of the classifier that leads to the so-called contextual classification.

Let us now proceed with a brief description of the classification method developed in this paper. The method is based on two assumptions resulting from practical observations:
\begin{enumerate}
\item A distinction must be made between two concepts specific to the human controller: the user's intention to perform the movement of the prosthesis and the user's imagination of the movement of the lost hand.
The first concept is a thought -  a plan to use the prosthesis for the intended purpose. The second concept is a physical quantity in the form of a nerve signal that activates the muscles of the stump, which in turn is the basis for generating the sEMG signal.

\item Usually, the operation of a dexterous (multifunctional) bioprosthesis is a sequence of movements performed in logical order and composing the desired (expected) action of the prosthesis. Typically, such a sequence consists of manipulating the empty prosthesis for its proper positioning relative to the gripped object  (including pre-shaping),  gripping the object, and manipulating the gripped object.

\end{enumerate}

The consequence of the first assumption is the distinction in the recognition system between the class (imagination of movement) of the object represented by the sEMG signal, and its interpretation (intention of movement) or actual movement performed by the prosthesis. In turn, the second assumption results in the division of the decision space covering all actions of the prosthesis, i.e., all sequences of movements into separate areas called boxes.
The boxes are clearly defined by events related to the performance of specific movements by the prosthesis.
Each box has a different classifier focused on the local recognition task, and also its own interpretation of the recognition result. As a result, we get a multiclassifier system, which is called a context-dependent recognition system due to its dependence on the events (context) that define the interpretation of the classification results.
The proposed structure of a context-dependent recognition system allows two goals to be achieved: (1) increasing the repertoire of movements performed by the prosthesis beyond the number of classes of sEMG biosignals; (2) improving the quality of classification through optimising the operation of the local (in boxes) classifiers of the multiclasifier system.

To the best of the authors' knowledge, this work is the first to propose a complex recognition method based on a multi-classifier system (MCS) operating according to a sequential context scheme. It is used to control a bionic hand prosthesis and allows for an increase in the repertoire of controlled movements (grasping and manipulation) of the prosthesis. The structure of the proposed MCS system can be optimised to achieve higher classification quality. Additionally, it should be emphasised that the action of the prosthesis user is included in the framework of motor activation of the residual limb muscles and is not limited by any restrictions, but depends only on the user's imagination.

The works known in the literature concern the use of individual methodological aspects for the control of a bionic prosthesis and the analysis of biosignals (in particular sEMG), only separately. On the other hand, the developed method comprehensively covers all of the above-mentioned aspects.

The most important aspects of this paper are as follows:
\begin{itemize}
    \item A proposal for a myoelectric control system of the bioprosthesis of the upper limb, in which two concepts were distinguished: (1) the user’s intention to perform the movement of the prosthesis; (2) the user’s imagination of the movement of the lost hand. Thanks to this, the pattern recognition paradigm with a context-dependent multiclassifier system can be used for control.

    \item The development of the structure of a context-dependent multiclassifier system that allows two goals to be achieved: (1) increasing the repertoire of movements performed by the prosthesis beyond the number of classes of sEMG biosignals; (2) improving the quality of classification by optimising the operation of the local classifiers of the multiclasifier system.

    \item The formulation of two optimisation problems related to the quality of the context-dependent classification system as combinatorial problems with constraints. The proposal of two solution procedures: a procedure based on an exhaustive search, and a procedure using an evolutionary algorithm for a permutation representation of individuals.

    \item The conducting of comprehensive experimental studies based on real signals from 10 amputees with a transradial amputation, and also from an able-body subject with an immobilised hand simulating an amputation. An original testing strategy was adopted in the research, taking into account the specificity of the considered recognition task, with two new quality measures resulting from this scheme then being applied.

\end{itemize}

The main research questions that are to be answered are as follows:
\begin{itemize}
    \item Does the context-dependent classifier achieve a better or similar classification quality to the context-free classifier?

    \item Does the box-structure optimisation (selecting proper movement-to-class binding) allow the context-dependent method to perform better when compared to a random (among the feasible solutions) box-structure choice?
\end{itemize}

The remainder of the article is organised as follows.
In Section~\ref{sec:RelWorks} we present works related to the issue discussed in this paper.
In Section~\ref{sec:systemofhandbioprosthesiscontrol} we start with a brief description of the myoelectric control system of the upper limb bioprosthesis, the structure and operation of which are fully compatible with the computer control system known from the field of automatic control. Here, we expose the elements of the system that are of key importance for the proposed method of the context-dependent interpretation of the results of the sEMG biosignal classification.
Section~\ref{sec:contextdeprecsystem} provides insight into the context-dependent recognition system as a decision-making  module of the hand bioprosthesis. First, the structure of the system is described, with its operation being illustrated with various practical examples, and then formal models of classification tasks, local classifiers, and the context-dependent system are presented. Two problems of optimisation of the context-dependent system are then formulated, and two methods of solving them based on the exhaustive search and the evolutionary algorithm are given.  
In Section~\ref{sec:experiments}, which concerns experimental research, we successively present sets of signals, the methods of the acquisition of the sets, the scheme (protocol) of comparative studies, the methods of extraction, the selection of features, and classifier models. The obtained results are presented in Section~\ref{sec:resultsanddisc}. The discussion is provided in Section~\ref{sec:discussion}. Finally, concluding remarks are presented in Section~\ref{sec:conclusions}.

\section{Related Works}\label{sec:RelWorks} 

In paper~\cite{Freitas2023} a simple MCS system with static selection of the best base classifier was used to recognise SIS (surgical instrument signaling) gestures. Paper~\cite{Kurzynski2017} presents an MCS system that uses a DES (dynamic ensemble selection) scheme and a customised competence measure. Well-known bagging and boosting schemas have also been used in tasks of controlling upper and lower limb prostheses~\cite{Akbulut2022}. Additionally, multiclassifier schemas were also utilised to recognise the hand gestures of able-bodied individuals~\cite{Subasi2020}. All of the methods presented in the above-mentioned papers are aimed at improving classification quality by using a multiclassifier system. In our work, on the other hand, the MCS uses a DCS (dynamic classifier selection) schema with an original classifier selection method that takes the classification context into account. The aim of this DCS is to increase the number of gestures and hand manipulations that can be used in a prosthesis. Furthermore, the proposed DCS procedure can improve the overall classification quality of MCS.

Increased prosthesis dexterity can also be achieved using targeted muscle renervation (TMR) surgery. In~\cite{Simon2023}, a case study is presented investigating this procedure. An increase in prosthesis dexterity may also be achieved using electroneurography (ENG). ENG signals are obtained from peripheral nerves by microelectrodes implanted in residual nerves~\cite{Nguyen2020}. A similar effect may be achieved using kinetico-myographic signals (KMG)~\cite{Moradi2022}. This kind of signal is generated by magnetic tags surgically implanted in the tendons. The authors prove that the SNR ratio of KMG signals is better compared to sEMG signals. All the methods presented above are invasive and do not always achieve success. There are many conditions that prevent them from carrying out the procedure~\cite{Rask2023}. Unlike them, the system proposed in this paper is not invasive and is not restricted by medical conditions.

The paper~\cite{Schone2023} presents the results of experimental research in biomimetic and non-biomimetic control strategy of the prosthesis. The biomimetic schema emulates the biological control of the hand by linking the imaginary movement of the phantom hand with the movement of the prosthesis. The non-biomimetic method allows arbitrary linkage between phantom limb movement and prosthesis action. We use non-biomimetic control in the context-dependent control schema presented in this paper.

A finite state machine (FSM) can be used to describe the proposed context-dependent control schema. Thus, it is necessary to mention works that use FSMs to describe the behaviour of the prosthesis. The main difference between the methods lies in the way in which the state change of the FSM is triggered. Transitions may be activated using a kind of external signal source such as a mobile application~\cite{Fajardo2021}, webcam~\cite{Fajardo2018}, or eye tracking~\cite{Shi2023}. The state may also be changed depending on the readings of sensors such as gyroscopes~\cite{Patel2017} or goniometers~\cite{Batzianoulis2019}. Invoking the change can also be done using biosignals such as sEMG~\cite{Nacpil2019}, mechanomiographic signal (MMG)~\cite{Geng2012}. Biosignal control can use a simple thresholding strategy~\cite{Nacpil2019} or machine-learning-based approaches~\cite{Cardona2020}. The transition between states may be discrete~\cite{DAccolti2023} or continuous~\cite{Piazza2020}. The presented context recognition system, interpreted as an FSM system, is fully automatic, based solely on the multi-channel sEMG signal, and does not use any additional information (signals, images). A change in the system state, meaning a change in the context determining the interpretation of classes and the active classifier model, is the result of the operation of the base classifier of the multi-classifier system. This is an innovative approach that allows achieving the expected benefits (improving the quality of classification, increasing the range of prosthesis movements).

\section{System of Hand Bioprosthesis Control}\label{sec:systemofhandbioprosthesiscontrol}

\begin{figure}[!hbt]
	\centering
		\includegraphics[width=0.95\textwidth, height=0.25\textheight, keepaspectratio]{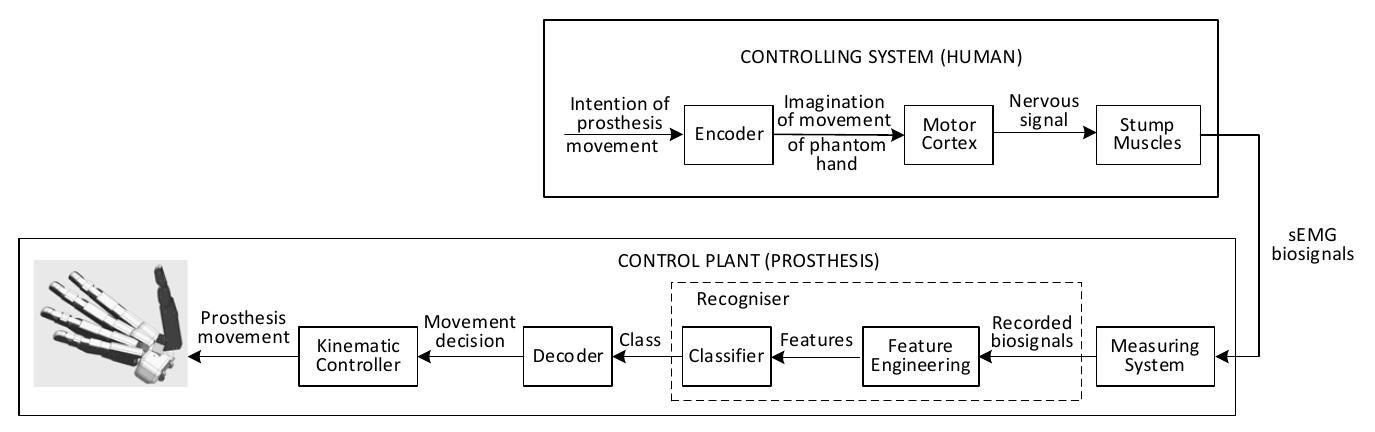}
	\caption{Open-loop control system for the bioprosthesis of the upper limb}
	\label{figs:BCS}
\end{figure}

Figure~\ref{figs:BCS} shows the block-diagram of an open-loop system for control a bionic multifunctional hand prosthesis based on the sEMG pattern recognition scheme. In this chapter, we will analyse the operation of the system from an IT perspective and introduce a formalism and terms  related to control theory, i.e.  we will present the key stages of information processing and the elements (modules) of the control system from Fig. \ref{figs:BCS}. Above all, however, we will highlight those features of the system that justify the main result of the paper, i.e., the context-dependent understanding (meaning) of the control decision, and also those system elements (modules) that show the location of this contextual interpretation in the entire process of the myoelectric control of a bionic hand prosthesis.

The action of the system starts when, in the controlling system, the intention of prosthesis movement is created. The intention as an input variable determines the control goal, which specifies the requirements for the controlled variable, i.e. prosthesis movement.
Next, a control algorithm is executed in the user’s mind, the result of which is a control decision. It provides information on how to control the object to achieve the goal. 
The term ''control algorithm''  is mainly associated with the technical nature of the controller.  In the control system considered with a human controller, this is the rule that the user uses to encode the intention of moving the prosthesis into contractions of the stump muscles. 
The control decision is in the form of a nerve signal that is generated by the motor cortex and transmitted to the residual muscles of the stump. These muscles, being a biological amplifier of nerve activity, play the role of the control actuator and convert the control decision into an EMG signal, which can then be recorded from the body surface (sEMG) and transferred to the input of the control plant as a control variable. Additional attention should be paid to the fact that the muscles of the stump change their properties over time (e.g. fatigue, change in fibre geometry, psychological factors, etc.) \cite{Farina2014}, causing the same control decision (neural signal) to have a different effect depending, for example, on the time of day. This means that the stump muscles are a non-stationary dynamic object. 

In turn, the control plant is entirely technical in nature. The purpose of the measurement system is to record the sEMG signals (after A/D conversion) that are generated by a human (controlling system) and further processed in the prosthesis, therefore, it can be called the human-prosthesis  (sEMG) interface. This interface should have a number of features that facilitate stable and repeatable EMG recording (secure sensor mounting, a fixed position relative to muscle fibres, a high signal-to-noise ratio, resistance to changing physiological conditions, negligible crosstalk between electrodes) \cite{Farina2014}.
Since we are considering a control system that uses the pattern recognition scheme, the main part of the control plant is a complex decision-making system called the recogniser. Within it, subsequent stages of the processing and analysis of the registered sEMG control signal are carried out, such as preprocessing (filtering, windowing), feature extraction (e.g. using discrete wavelet transformation (DWT)), feature dimensionality reduction (collectively called feature engineering), and classification. It should be noted that in the system the classifier usually operates under uncertainty conditions, which means that the features of the recognised object do not always provide unambiguous information about its class label. This is due to the non-stationary nature of the signal source (muscles of the stump), imperfections of the human-prosthesis interface, as well as the often poor discriminating power of human-generated sEMG signals \cite{Kristoffersen2021}. This means that the possibility of making errors by the classifier is inherently inscribed in the principle of its operation, with the efforts of designers that use increasingly sophisticated methods and algorithms \cite{Parajuli2019}, \cite{Chen2023} being to minimise the error rate.
The task of the kinematic controller is to control the DoFs of the prosthesis in such a way that the movement of the mechanical structure (trajectory of movement) ends with the posture of the prosthesis (grip or manipulation) being consistent with the classification result.

We now turn to those elements of the control system that are of key importance for the proposed method of the context-dependent interpretation of the results of the sEMG biosignal classification. These elements are the following two triads: intention -- encoder -- imagination in the controlling system, and class label -- decoder -- movement decision in the control plant.

The intention of the user is not an accomplished form, but only an idea or plan to use the prosthesis for the intended purpose. This intended purpose determines the type of movement the prosthesis should make or the posture it should take. In order to translate this intention into action, the user must activate the stump muscles according to some rule, i.e. execute a control algorithm. The user has full freedom to shape this rule, but the user is also aware that the effect of their effort (sEMG signal) should help the prosthesis classifier (to the maximum extent) in making the correct decision. At the same time, the user would like to achieve this effect intuitively. 
This rule seems simple in the case where the sEMG signal is obtained from an able-bodied person. When we want, for example, the prosthesis or robotic arm to make a cylindrical grip, we make this grip with the hand from which the signal is registered. Because in this automatic and fully intuitive operation the biological hand mimics the required movement of the bionic prosthesis, this method of control is called a biomimetic control strategy~\cite{Schone2023}.

The situation is not so obvious in the case of transradial amputations when the real hand is replaced by a phantom hand. The prosthetic control rule (algorithm) can then be based on the paradigm of phantom motor execution (PME), which is the phantom movement of the amputated limb that the user is able to voluntarily control \cite{Akbulut2022}. Voluntary PME has recently been shown to be a form of ''real'' motor execution, therefore activating stump muscles and being the source of sEMG signals \cite{Garbarini2018}.
When the muscles of the stump are activated, the sEMG signal is generated. It cannot be said that this was due to a specific movement of the hand (e.g. cylindrical grip) in accordance with the intention of the user. The activation of muscles can only be induced by imagining the movement of a phantom hand that does not exist. Given the intuitiveness of the procedure, this representation (image) of the phantom hand's movement should, ideally, be consistent with the desired movement of the prosthesis as defined by the user's intention (biomimetic control strategy).
However, it does not have to be that way, because the muscle contraction patterns of the stump during PME are different from those of the intact limb due to surgical intervention during amputation and movement of the amputated limb cannot be observed. Hence, it is difficult to expect the required movement of the prosthesis to be imitated by the phantom hand. Rather, it will be a non-biomimetic (or arbitrary) control strategy in which the user freely encodes the intention to move the prosthesis into the activation of residual muscles in the limb, i.e., executes the control algorithm.

From the above, it can be seen that the sEMG biosignal generated by the activation of the muscles of the stump is not directly related to the user's intention of making a movement by the prosthesis, as is commonly assumed in the literature. This sEMG signal is directly related to the imagination of the phantom hand movement, i.e. a specific ''movement without a movement''. This observation requires a reorientation of the meaning of the class in the classification task used in the bioprosthesis control procedure with the pattern recognition paradigm. Now the class to which the object represented by the registered sEMG signals belongs is the user's imagination of the movement of the phantom hand.

This distinction between the concept of \emph{the user's intention to perform the movement of the prosthesis} and the concept of \emph{the user's imagination of the movement of the lost hand} is not just a matter of terminology. By using this distinction, the possibilities of building a suitable bioprosthesis control system can be expanded without destroying existing concepts and solutions. We will demonstrate this through various modes of operation and formal descriptions of the encoder and decoder.

The relationship between the intention ($In$) and the image ($Im$) must be described by a bijective function (one-to-one and onto function). On the one hand, the user must clearly know how to translate his intention $In$ into his imagination $Im$ that generates the sEMG signal. On the other hand, the prosthesis must clearly ''know'' how to decode the classification result $Im$ and obtain the operation of the kinematic control system that leads to the movement of the prosthesis according to the intention $In$.
Possible cases of encoder and decoder operation, defining 3 levels of relationships between sets $\mathscr{I}m$ and $\mathscr{I}n$, are presented in Table \ref{tab:EncoderDecoder}. Although the human control system includes advanced neuromuscular processes taking place in the user's body, when characterising the formalism of the encoder operation, we will limit ourselves to the effects of these processes important for the prosthesis control system from a technical perspective. Levels \rom{1} and \rom{2} describe a typical encoder which, within the phantom motor execution (PME) paradigm, maps the user's intention into the activity of the residual limb muscles, i.e., according to the adopted interpretation, into the image of a phantom movement. Both levels differ in the prosthesis control strategy and the nature of the mapping. For level \rom{1}, it is biomimetic motor control, in which the phantom hand mimics the desired bioprosthesis movement. That is, the user's intention and user imagination are the same. Level \rom{2} concerns a non-biomimetic (arbitrary) control strategy in which the imagination of moving the phantom hand is not identical to the desired movement of the prosthesis. For example, in the experimental studies presented in~\cite{Schone2023} 4 movements of the prosthesis (open position, closed position, pinch grasp, and tripoid grasp) were controlled by sEMG patterns obtained from the stump muscles activated by completely different movements of the phantom hand (extending one, two, three and four fingers, respectively). The user has full freedom to shape the $T$ function, that is, mapping the movements of the prosthesis into phantom movements. Level \rom{3} of the encoder operation can be treated as level \rom{2} with an additional state-switching mechanism that changes the $T$ function.

 The presented formalism with respect to the operation of the decoder is clear and understandable. It concerns the classifier-kinematic controller interface, which directly defines the relationship between the recognised class and the movement to be performed by the prosthesis. This relationship is either static (for levels \rom{1} and \rom{2}) or depends on additional circumstances called context (level \rom{3}), i.e., it has a dynamic nature. The class--movement relationships are established at the stage of building the control system in the supervised learning procedure and, for level \rom{3}, additionally by determining the structure of the context system. Both procedures are carried out with the active participation of the user. In creating the training set, it is the user who labels the recorded biosignals, simultaneously defining the $\mathscr{I}m$ set. The user also determines the outcome produced by the kinematic control system. That is, the user determines the $\mathscr{I}n$ set consisting of the prosthesis movements that are most useful in one’s everyday life.

Thus, the presented formalism is a user-oriented concept and guarantees consistency in the operation of the encoder and decoder. Level \rom{3} of encoder/decoder operation is utilised in the contextual classification method proposed in the paper. In the next section, we will describe the developed method in detail and unambiguously define the concept of context i.e., a switch between class meanings, which allows one to increase the number of prosthetic movements that are possible to control without increasing the number of classes in the recognition task.

\begin{table}
\centering
\footnotesize
\caption{Three levels of encoder and decoder operations}\label{tab:EncoderDecoder}
\begin{tabular}{|c |c |c |l |}
\hline
Level & Encoder & Decoder & Comment \\
\hline
\rom{1} & $Im=In$ &  $In=Im$ & Identity function.  Includes cases where no distinction  \\
&  & & is made between $In$ and $Im$.\\
\hline
\rom{2} & $Im=T(In)$ & $In=T^{-1}(Im)$    & $T$ – a bijective function. Bijection $T$ can be conveniently \\
&  $Im \in \mathscr{I}m$, $In \in  \mathscr{I}n$ & &  described using the LUT (Look Up Table) method. \\ 
\hline
\rom{3} &  $Im=T_p(In)$  & $In=T_p^{-1}(Im)$    &  $p$ – a discrete parametr called context. All $T_p$ \\
& $Im \in \mathscr{I}m_p$, $In \in  \mathscr{I}n_p$& & must be bijective functions.  \\
& & &  $\cup_p\mathscr{I}m_p = \mathscr{I}m$ and $\cup_p\mathscr{I}n_p = \mathscr{I}n$. \\
\hline
\end{tabular}
\end{table}

The operation of the encoder and decoder at the \rom{2} and \rom{3} level of relationship can be difficult for users to accept because of poor intuitiveness. Intuitiveness is a measure of the discrepancy between performing some activities to the best of the user's abilities and performing them according to expectations. This difference can be reduced, and thus intuitiveness can be increased by reducing expectations or by increasing the user's capabilities. The latter effect is obtained by training (acquiring knowledge) \cite{Kristoffersen2021} or a modified (enriched) user interface. In \cite{Kurzynski2017a}, the authors – based on brain cortex plasticity – propose the computer-aided training system, which when generating visual (via virtual reality) and sensory stimuli should enhance the effectiveness of mental training of the control of voluntary movements. In turn, the paper \cite{Dyson2020} presents and experimentally examines motor learning-based methods in which the patterns of muscle activity used for prosthetic control can differ from those which control biological limb. Thus, an originally non-intuitive solution can become intuitive and effective after implementing appropriate training activities for the amputee.

\section{Context-dependent Recognition System}\label{sec:contextdeprecsystem}

\subsection{Fundamentals}\label{sec:contextdeprecsystem:fundamentals}

The method proposed in the paper is related to the context-dependent interpretation of the recognition result, which is not an original idea. In image recognition, the problem of the interpretation of objects and their configuration has to be solved in different applications, such as medical imaging (diagnosis), autonomous mobile systems, or remote sensing. Any system for the interpretation of images uses a priori knowledge of the origin and properties of the image, the actions visible in the image, and the conclusions resulting from the content of the image. The application of semantic or statistical models for the formal representation of activities in image interpretation leads to the novel cognitive methodology beyond recognition called image understanding \cite{Sarvamangala2021}. The concept of understanding recognition results was significantly developed in the book \cite{Grzegorzek2017}, which presents the interpretation of human data that describe physical activity, cognitive activity, and emotion recognition. 
 
In the proposed scheme of the control system, the classification of sEMG signals does not end the operation of the decision-making part of the control plant. The recognised class is subject to additional interpretation in the decoder, and only the result of this interpretation indicates the movement that the prosthesis should perform. Since this interpretation depends on the context of classifier operation, the entire decision-making system will be briefly called a \textbf{context-dependent recognition system}.

Context in recognition problems can take many forms. In our task, this context will be determined by the result of the previous classification of another classifier. A specific structural order in the bioprosthesis control system at the decision-making (classification) level allows for the convenient determination of both the event that unambiguously defines a given interpretation of the classification results and the event that ends this interpretation in favour of another. This interval between the two events, where a particular interpretation of the classification results is valid, will be called \textbf{the box}. Each box runs a different classifier used in the local recognition task, and each box has a local interpretation of the classification results.
The proposed structure of isolated boxes allows two goals to be achieved: (1) increasing the repertoire of movements performed by the prosthesis beyond the limit resulting from the number of generated patterns (classes) of biosignals, (2) optimising the operation of local classifiers by selecting their models, features, recording channels, and interpreting classes adopted in a given box.

We will now present, in a descriptive form illustrated with examples, the contextual recognition system and its box structure. 

First, the repertoire of movements (grasping or manipulating) that the prosthesis can perform must be determined. Then we create sequences of movements that, in the opinion of the potential user of the prosthesis, are of practical importance and will be useful in everyday life. A natural sequence of movements can be created by manipulating the empty prosthesis, positioning it relative to the object, gripping the object, and then manipulating it. If there is a pair of movements in such a sequence, the first of which is to perform a movement (grip or manipulation), and the second is to return the prosthesis to its original position, then we are then dealing with a box. We say that the first movement of the pair opens (initiates) the box and the second closes (ends) the box. For the box not to be empty, there must be movements between its opening and closing. Therefore, the box is opened when, as a result of classification, the prosthesis performs a certain movement. It is closed when, also as a result of classification, the same movement is completed (the prosthesis returns to its pre-movement state). The movements between these events form the inside of the box.

Let us imagine, for example, that one of the planned sequences of movements is the use of a computer mouse and the pronation position is the initial state of the prosthesis. Therefore, we have the following sequence of movements:
\begin{enumerate}
    \item Start prosthesis position: pronation;
    \item Box opening: mouse grip;
    \item Movements inside the box: momentary index/ring finger flexion (in any order and number);
    \item Box closing: mouse grip release;
    \item Final prosthesis position: pronation. 
\end{enumerate}

In the example, the box was opened (initialised) as a result of recognising the class interpreted as the mouse grip. This recognition occurred before the box was opened, and therefore the classification had to be made by another classifier operating earlier (e.g. in another box). Inside the box, there is a classification task that includes 3 classes interpreted as the following movements: (1) momentary index finger flexion, (2) momentary ring finger flexion, (3) mouse grip release. The last recognition closes the box. Any classifier operating inside a box will be called a box classifier. 

Note that when talking about the classifier inside the box, we did not specify the classes (imagination of phantom hand movements) that it recognises, but only the movements (intention of prosthesis movements) that interpret the classes. Therefore, in order for the classification task in a box to be unambiguously defined, we must give the classes a local (inside a given box) interpretation, or, due to the bijective function associating movements with classes, assign movements to classes. By emphasising the reflexive nature of both relationships, we will denote this interpretation (or connection) by $\mathrm{Movement} \leftrightarrow \mathrm{Class}$.

The boxes that create the recognition system can be positioned in relation to each other in three ways: (1) A nesting relationship is when we open one box, then open another one inside it, and then close the second box earlier than the first one. In other words, the second box is nested inside the first box. The nesting level will be called the order of the box. The same order is assigned to the box classifier. (2) A serial relationship is when we open one box, then close it, then open the other, and close it too. Both boxes occur in the same sequence of movements, so both must be executed. (3) A parallel relationship occurs when the boxes are located in different sequences of movements and cannot be performed on the same action.

\begin{figure}[!htb]
	\centering
		\includegraphics[width=0.8\textwidth, height=0.45\textheight, keepaspectratio]{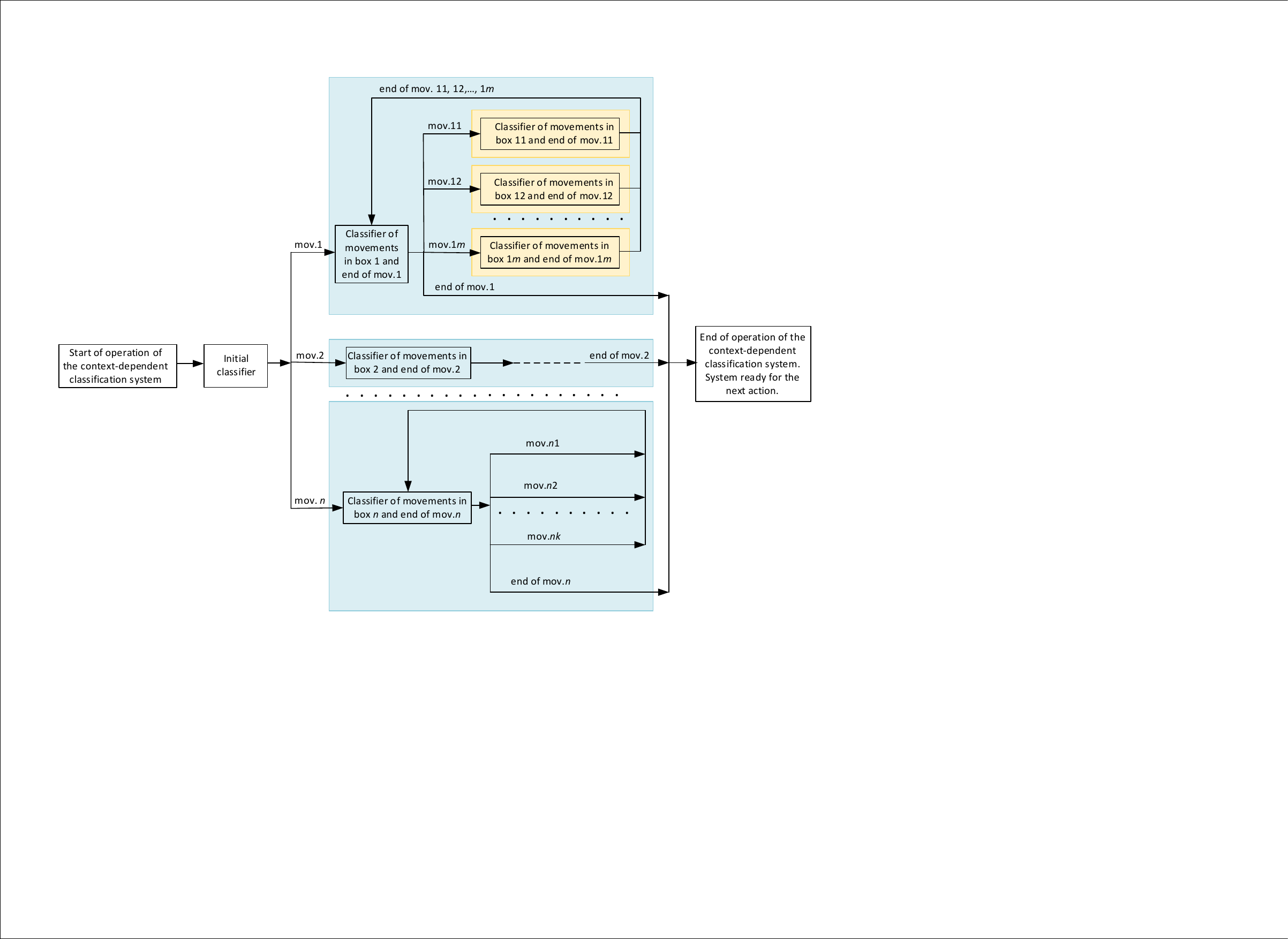}
	\caption{An example of the structure of the context-dependent classification system}
	\label{figs:Box1}
\end{figure}

Fig.~\ref{figs:Box1} shows an example of the context-dependent classification system and its box structure. The movements in the figure are not assigned any practical meaning, because only the presented decision scheme is important here. Thus, box $n$ demonstrates an example of possible classification actions, and box $1$ illustrates nested (yellow) boxes. The operation of the context-dependent classification system starts with the so-called initial classifier or zero-order classifier. This classifier is located in the zero-order box that is opened automatically. This is clear because there is no other classifier before the initial classifier in which the classification result could open the zero-order box.

\begin{figure}[!htb]
	\centering
		\includegraphics[width=0.6\textwidth,height=0.33\textheight, keepaspectratio]{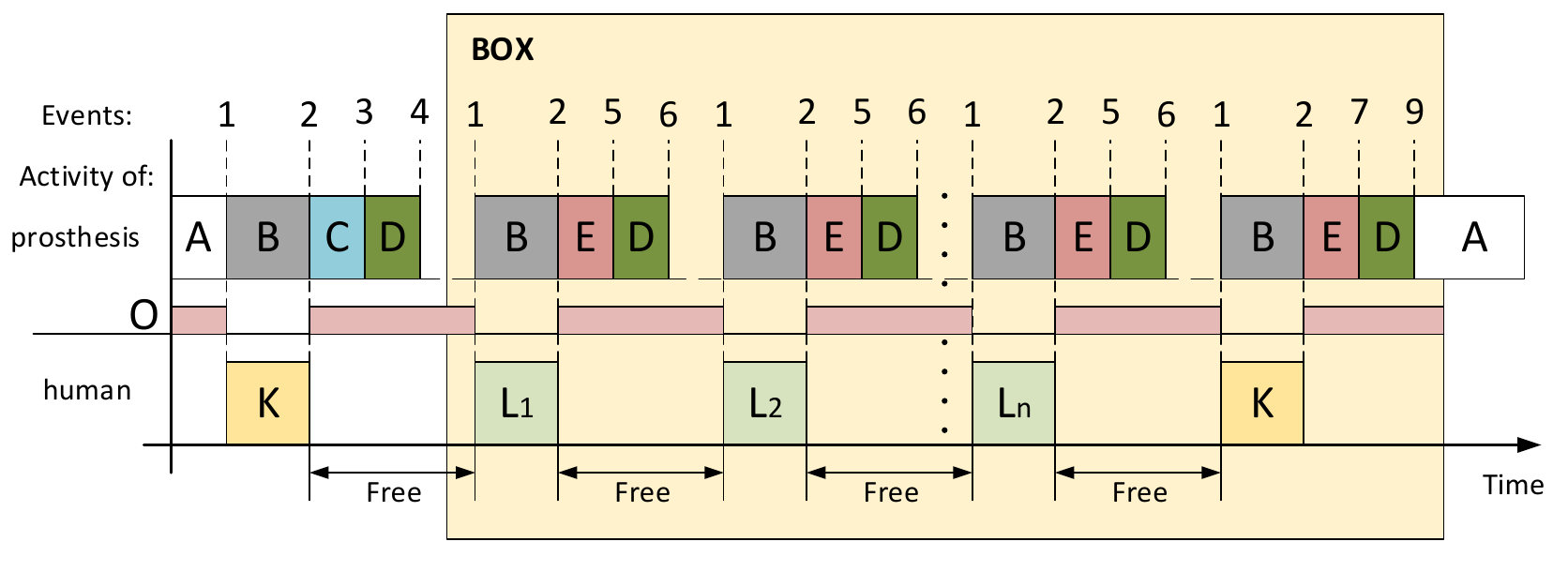} 
	\caption{Time scheme of human and prosthesis activity inside a box (for the sake of simplicity,  all classification results are referred to their interpretation, i.e. to the intention of the movement).	
	Events: (1) - onset detection; (2) - offset detection or timeout $T$; (3) - recognition of the intent to perform the box-opening  movement; (4) - performing  recognised box-opening movement; (5) - recognition of the intent to perform a movement inside the box; (6) - performing a recognised movement inside the box; (7) - recognition of the intent to perform the box-closing  movement; (9) - performing a recognised box-closing movement; bringing the prosthesis to its initial state. 
Prosthesis activities: (A) - prosthesis in the initial position (state) - e.g. the resting position; (B) - recording of detected sEMG biosignals; (C) - operation of the classifier external to the box; (D) - operation of the kinematic control algorithm; (E) - operation of the classifier inside the box; (O) - the prosthesis is listening. 
Human activities: (K) - generating sEMG biosignals indicating the intent to perform the box initiating/closing movement; (Li) - generating sEMG biosignals indicating the intent to perform subsequent movements inside the box. 
}
	\label{figs:Time}
\end{figure}

The initial classifier and box classifiers recognise classes in which interpretation in the form of movements determines the operation of kinematic control algorithms, and in turn causes appropriate actions of the prosthesis. 
The comprehensive set of events that occur in the example box, as well as the interactive actions of the prosthesis and the human are presented in Fig.\ref{figs:Time}.

\begin{figure}[!htb]
	\centering
		\includegraphics[width=0.40\textwidth]{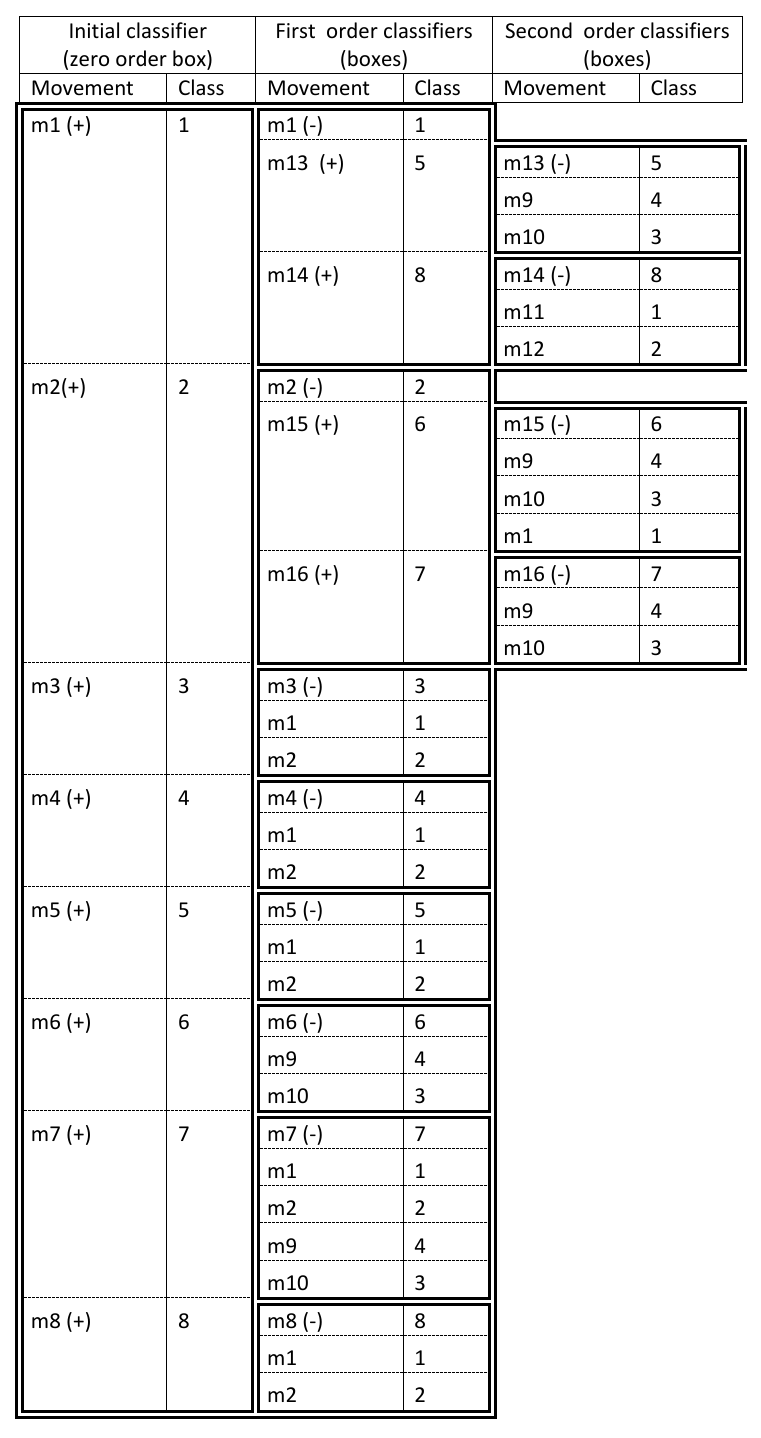}
	\caption{Example of the structure of the context-dependent classification system. Types of movements \cite{Cini2019}: (m1) pronation; (m2) supination; (m3) oblique grip; (m4) hook grip; (m5) spherical grip; (m6) cylindrical grip; (m7) precision grip; (m8) key grip; (m9) wrist flexion; (m10) wrist extension; (m11) index finger flexion; (m12) ring index flexion; (m13) finger point; (m14) mouse grip; (m15) lateral grip; (m16) platform grip. The symbol $(+)$/$(-)$ means that the movement initiates/closes the box. Boxes are marked with a double line.}  
		\label{figs:UR}
\end{figure}

Fig.~\ref{figs:UR} shows a context-dependent classification system in which movements have a specific practical meaning and have been associated with classes according to the optimisation algorithm 2 described in Section~\ref{sec:contextdeprecsystem:optimization}.  The structure of the classifier and the layout of the boxes result from the adopted scheme of the sequence of movements creating the possible actions of the prosthesis. The scheme is related to experimental studies conducted on an able-body person with an immobilised hand, for which 8 classes have been identified (see the description in Section~\ref{sec:experiments}). By taking advantage of the possibility of various interpretations of classes in the structure of the classifier, the actions of the prosthesis involving 16 movements were proposed. This means that each class has two interpretations. A precisely defined context (box) means that both interpretations are always unambiguous, which is clearly visible in the analysis of columns ''Movement'' and ''Class''.

\subsection{Model}\label{sec:contextdeprecsystem:model}

We will now present the general formalism of the classification problem, which is the paradigm of the operation of the control plant (upper limb bioprosthesis) of the control system presented in Fig.~\ref{figs:BCS}. The contextual approach will not be taken into account for now. As we already know, classes denote the image of the movement of the phantom hand. Classes therefore have - at the classification level - a specific meaning known only to the user, and do not have to (although they may) coincide with the meaning of the movements performed by the prosthesis. This meaning will be determined at the stage of the interpretation of the classification result. From the point of view of the recognition task, the meaning of the classes is irrelevant. Therefore, we do not have to associate verbal descriptions with them, but we will label the classes with consecutive natural numbers. Let then:
\begin{equation}   \label{1}
 \mathcal{C}=\{1, 2, \cdots, C\}
\end{equation}
denote a set of class numbers (labels). The number of classes $C$ for each user may be different, as it is related to the size of the PME repertoire (see Section~\ref{sec:systemofhandbioprosthesiscontrol}) and its relation to the activation of the stump muscles.

Since we are dealing with the bionic control system, the carrier of information about the class of the recognised object are sEMG biosignals. 
After extracting features from biosignals, the representation of the recognized object is a vector of features: 
	\begin{equation}     \label{2}
	 x =(x^{(1)}, x^{(2)}, \cdots, x^{(d)}) \in \mathcal{X} \subseteq \mathbb{R}^d,
\end{equation}
belonging to the $d$ dimensional feature space $\mathcal{X}$. 
The methods and techniques for extracting time series features are very well known \cite{Barandas2020}. Many of them have been used in the bioprosthesis control task with varying success \cite{MendesJunior2020}. In experimental studies (presented in Section~\ref{sec:experiments}), the individual-sEMG-biosignal-recording-channel features were extracted using the discrete wavelet transform (DWT) \cite{Shariatzadeh2023}.

In practical recognition problems, an additional dimensionality reduction procedure for object representation is usually used in the form of feature selection \cite{Khan2023}, or the feature reduction method \cite{Velliangiri2019}. Let:
\begin{equation}        \label{3}
\bar{x} \in \bar{\mathcal{X}} \subseteq \mathbb{R}^{\bar{d}}
\end{equation}
be the feature vector after the application of the feature selection/reduction procedure ($\bar{d} < d$).
We will now specify the classifier for the recognition task defined by formulas \eqref{1} and \eqref{3} as a function:
\begin{equation}  \label{4}
\psi(\bar{x})=j; \; \;  \bar{x} \in \bar{\mathcal{X}}, \; \; j \in \mathcal{C},
\end{equation}
which maps the feature space after dimensionality reduction into a set of class numbers, namely:
\begin{equation}   \label{5}
 \psi: \; \;  \bar{\mathcal{X}} \rightarrow \mathcal{C}.
\end{equation}
We assume that the classifier \eqref{5} is built in the supervised learning procedure, i.e., using the training set $\mathcal{N}_u$ containing $N$ labelled objects from the feature space $\bar{\mathcal{X}}$.

The upper limb bioprosthesis, in accordance with the design assumptions and its degrees of freedom (DoF), is capable of performing the following set of movements (grips and manipulations):
\begin{equation}  \label{6}
\mathcal{M}= \{m_1, m_2, \cdots, m_M\}.
\end{equation}
If the bioprosthesis is to be multifunctional (dexterous), the required number of $M$ movements should be considerable. For example, taking into account the six wrist manipulations (flexion--extension, pronation--supination and ulnar--radial deviation) and the grasps collection according to the taxonomy and division into 3 types (power, intermediate, and precision grasps) \cite{Cini2019}, then the number of movements performed by the prosthesis should be $M=26$. Achieving a multifunctional mechanical construction of the prosthesis with multiple DoFs is currently not a problem \cite{Calado2019}. The problem is the difficulty in controlling such a number of movements, because in the pattern recognition method, the number of classes $C$ usually does not exceed 12 \cite{Mendez2021}. Therefore, it is not possible to fully control the multifunctional hand prosthesis based on the classical pattern recognition system due to the impassable limit $M\leq C$.  
This problem can be mitigated by the proposed context-dependent classification system. The developed method allows for giving classes many interpretations that remain unambiguous, which is due to the separation of contexts (boxes) determining this interpretation.

Before proceeding to the formal description of the context-dependent recognition system as a multi-classifier system containing an initial classifier and box classifiers, we will present the assumptions along with the limitations. Not all restrictions are mandatory. Some assumptions are critical in order to form a contextual approach, while the others arise from the will to give classes (imaginary movements of the lost hand) different meanings (prosthesis movements) that are as intuitive as possible. Thus, the limitations that were adopted are a tribute to the user's convenience while using the prosthesis, and take into account the actual potential for activating the stump muscles. 

Assumptions:
\begin{enumerate}
\item  Each class from the set \eqref{1} has a primary meaning (movement from the set \eqref{6}). Although this attribution of primary meaning can be made freely, we assume that it will not be subject to any formal action (e.g. it will not be an optimising variable in the optimisation tasks presented in the next subsection). We assume that this assignment of the primary meaning is arbitrary or results from the best possible intuitive connection of the class with the movement of the prosthesis.  Without the loss of generality, we can assume the following primary meanings of classes:
\begin{equation}   \label{7}
\mathrm{class} 1 \leftrightarrow m_1,\; \; \mathrm{class} 2 \leftrightarrow m_2, \; \cdots, \mathrm{class} C \leftrightarrow m_C. 
\end{equation}
We also make the pragmatic assumption that all classes in \eqref{1} have a primary meaning (interpretation).

\item Each movement from the set \eqref{6} can define (initiate, open) a box. In such a case, the movement initialising (opening) the box and the movement closing the box are connected to the same class. This principle should make it easier for the user to navigate the structure of the boxes, because entering and leaving the box requires the same activation of the stump muscles, which seems to be quite an intuitive solution. 

\item We assume that the initial recognition task covers all $C$  classes from the set \eqref{1}, and that the classes are assigned a primary meaning \eqref{7}. The first step of the prosthesis action should have the widest range of possible activities, because it determines the number of sequences of movements performed by the prosthesis, that is, actions on which the dexterity (multifunctionality) of the prosthesis depends on.

\end{enumerate}

The context-dependent classifier is a multi-classifier system, as it consists of classifiers dedicated to recognition tasks in individual boxes, and an initiating classifier. Due to assumption 3, the initial recognition task is identical to the general classification task described by formulas \eqref{1} and \eqref{3}, with the initial classifier of form \eqref{4} and \eqref{5} being trained using the learning set $\mathcal{N}_u$. The difference between a context-free and a context-dependent classification appears only at the box level greater than zero.

Let the $l\mathrm{-th}$ box in the context-dependent classifier structure include the following movements:
 \begin{equation}   \label{8}
m_{l}^{(1)} (\textrm {box-closing movement}),  m_l^{(2)}, \cdots, m_l^{(M_l)} (\textrm {internal movements in the box}), \;  l=1,2,\cdots,L,
\end{equation}
where $L$ denotes the number of boxes, and $M_l$ is the number of movements in the $l\mathrm{-th}$ box (including the movement that closes the box). Note that based on assumption 2, movement $m_l^{(1)}$ already has a class assigned to it, which will be denoted by $j_l \in \mathcal{C}$. Thus, in the $l\mathrm{-th}$ box we have a classification task with a number of classes equal to $M_l$, among which is the class $j_l$. It is obvious that $M_l \leq C$.
We will now formally describe the classification problem in the $l\mathrm{-th}$ box ($l=1,2,\cdots ,L$) in a similar way to the general classification problem \eqref{1} and \eqref{3}:
\begin{enumerate}
\item { The set of class numbers for the $l\mathrm{-th}$ box:
\begin{equation}   \label{9}
 \mathcal{C}_l=\{j_l^{(1)}=j_l, j_l^{(2)} , \cdots, j_l^{(M_l)}\}, \; \;  \mathcal{C}_l \subseteq \mathcal{C}. 
\end{equation}
The set of classes \eqref{9}, which defines the recognition problem in the $l\mathrm{-th}$ box, depends on the way of giving the classes a secondary interpretation in terms of movements \eqref{8} that occur in this box. This interpretation (or the association of movements with classes) may be subject to various constraints, e.g., the reserved class $j_l$ associated with the box-closing movement (this will be discussed in more detail in the next subsection). Let us denote by $\mathscr{C}_l$ the family of sets \eqref{9}, i.e., sets of class numbers derived from \eqref{1} with cardinality $M_l$, taking into account possible constraints. In other words, $\mathscr{C}_l$ is a family of possible sets of classes that can be associated with movements in the $l\mathrm{-th}$ box}. Thus: 
\begin{equation}  \label{9a}
{C}_l \in \mathscr{C}_l, \;  \;   l=1,2, \cdots, L.
\end{equation}
 \item {The feature vector after the selection/reduction procedure, which is the basis for the operation of the classifier in the $l\mathrm{-th}$ box:
\begin{equation}        \label{10}
\bar{x}_l \in \bar{\mathcal{X}_l}\subseteq \mathbb{R}^{\bar{d}_l}, \; \; \bar{d}_l < d.  
\end{equation}
The feature vector \eqref{10} depends on the size and composition of the set of classes \eqref{9} and is the result of subjecting the original feature vector \eqref{2}, i.e., the quantitative representation of the recognised object, to a specific procedure of reducing the dimensionality of this representation. It may be feature selection based on a filter scheme with a selected measure to assess the discriminatory power of individual features. Alternatively, a wrapper procedure can be used with an adopted optimisation method \cite{Kaur2022}. This can also be a feature reduction, in which a less dimensional feature space $\bar{\mathcal{X}_l}$ is obtained by transforming the original $\mathcal{X}$ space (e.g. the Principal Component Analysis (PCA) method) \cite{Rabin2020}. 
}
\end{enumerate}
The formal description of the classifier in the $l\mathrm{-th}$ box is as follows:
\begin{equation}  \label{11}
\psi_l(\bar{x}_l)=j; \; \;  \;  \bar{x}_l \in \bar{\mathcal{X}}_l, \; \;  \; j \in \mathcal{C}_l,
\end{equation}
or:
\begin{equation}   \label{12}
 \psi_l: \; \; \; \bar{\mathcal{X}_l} \rightarrow \mathcal{C}_l.
\end{equation}
As before, we assume that $ \psi_l$ is trained in a supervised learning procedure based on the learning set $\mathcal{N}_u^{(l)}$ containing $N_l$ objects from the set $\bar{\mathcal{X}_l}$ and their labels from the set \eqref{9}.

To fully define the recognition task in the $l\mathrm{-th}$ box, it is still necessary to explain how the set of classes \eqref{9} is determined, that is, how we associate classes with movements \eqref{8} in the $l\mathrm{-th}$ box. In other words, how do we choose the set $\mathcal{C}_l$ from the family $\mathscr{C}_l$, and how do we determine this family.
We will answer these questions in the next subsection.

\subsection{Optimisation}\label{sec:contextdeprecsystem:optimization}

Organising the structure of the context-dependent classification system into boxes, in which separate classification tasks are defined, opens up space for optimising the recognition system. After assigning classes to internal movements of the box, and taking into account the class related to the box-closing movement, we receive a clearly defined local classification task. Then, at the learning stage, we can tailor the classifier $\psi_l(\bar{x}_l)$ in terms of the model and the selected features (for the specificity of this task), which should positively affect the quality of the classification. At the same time, however, the quality of the classifier is influenced by the classes assigned to the movements in the box. This is a known effect in pattern recognition: if we have a classification task limited to a certain subset of classes, then the ''difficulty'' of the task is affected by the number of classes (the size of this subset) and the classes from which this subset is composed. This is because some classes are more easily discriminated than others \cite{Lorena2019}. In our problem, the size of the class set is the number of movements in the box (plus the box-closing movement), and the class numbers result from the way the movements are assigned. Therefore, it is possible to assign such classes to movements in order to obtain the best classification quality, i.e., the best discriminating set of classes $C_l$.

In the next part of the subsection, two optimisation problems will be presented and ways of solving them will be shown. One of the problems of practical importance will also be the subject of experimental research, which is described in Section~\ref{sec:experiments}.

\subsubsection{Optimisation problem I}\label{sec:contextdeprecsystem:optimization:one} 

The first optimisation problem is characterised by two features: (1) we treat each classifier in the system independently, which means that the optimisation of the entire context-dependent recognition system comes down to the optimisation of each classifier separately; (2) we do not impose any restrictions (except for the previously formulated assumptions) on the method of determining the interpretation of each class, nor on the number of these interpretations. This means that in the case of a large number of interpretations of the same class, the result may not have much practical significance, because it will be unacceptable to the user. However, we consider this case consciously, as it is interesting from an academic perspective and as a starting point for the second optimisation problem, which is already fully practical.

Optimisation problem 1, formulated for the $l\mathrm{-th}$ box ($l = 1, 2, \cdots ,L$) is as follows:

We are given: internal movements \eqref{8} and class number $j_l$; primary class interpretations \eqref{7}; primary feature space $\mathcal{X}$.

Goal: to find a set of classes $\mathcal{C}_l^* \in \mathscr{C}_l$ that maximises the criterion (objective function) $Q_l$, viz.
\begin{equation}   \label{14}
Q_l(\mathcal{C}_l^*) = \max_{\mathcal{C}_l \in \mathscr{C}_l} Q_l(\mathcal{C}_l). 
\end{equation} 
The problem \eqref{14} is a discrete and combinatorial optimisation problem, which can be expressed as a tuple ($\mathscr{C}_l, Q_l$).

\begin{enumerate}
\item The set $\mathscr{C}_l$ is a finite set of all feasible solutions (all sets $\mathcal{C}_l$) that satisfy the specified constraints. The movement $m_l^{(1)}$ in the set \eqref{8} that closes the box is already assigned the class $j_l$, which - according to assumption 2 - is the same as the class of the box-opening movement. This is the only mandatory constraint that defines the set (family) $\mathscr{C}_l$. Other existing assignments (interpretations) resulting from optimisation in other boxes are not taken into account. Therefore if we limit ourselves to this one condition, then \eqref{14} is a combinatorial optimisation problem in which the cardinality of the set of feasible solutions $\mathscr{C}_l$ is equal to the number of ($M_l - 1$)-combinations of the set of ($C-1$) classes, i.e.:
 \begin{equation}   \label{15}
\left| \mathscr{C}_l \right| = \binom{C-1}{M_l-1}=\frac{(C-1)!}{(M_l-1)! \; (C-M_l)!}.
\end{equation}

\item The objective function $Q_l$ is a function that maps each element of the set $\mathscr{C}_l$ to the set of real numbers. In the recognition task, the criterion should be related either to the quality of the classification (e.g. accuracy, kappa statistic, F-measure, AUC) \cite{Gong2021}, or to a measure evaluating the ''difficulty'' of the recognition problem \cite{Lorena2019}. In the first case, to determine the value of the criterion, we must have a reduced feature space \eqref{10}, a trained classifier \eqref{11}, and a test set, while in the second case only a training set $\mathcal{N}_u^{(l)}$ is needed. We do not treat the above-mentioned quantities needed to evaluate the optimisation result as optimisation variables, although they do affect the value of the criterion. It is important that the determination method of the criterion $Q_l(\mathcal{C}_l)$ for all solutions from $\mathscr{C}_l$ is the same and that the comparison of results is fair. In the experimental studies presented in Section~\ref{sec:experiments}, a scheme was adopted in which the classifier model, the feature selection/reduction method, and the learning and testing procedure (e.g., $k \times n$ cross validation method) were first determined, and then these models and methods were used for the calculation of the value of the criterion $Q_l$ for all evaluated solutions $\mathcal{C}_l \in \mathscr{C}_l$.

\end{enumerate} 

Note that the solution to the problem \eqref{14} is a set of class numbers that maximise the criterion $Q_l$, and it is not necessary (in the problem solving stage) to assign internal movements to the classes.
Although this assignment has no effect on the quality of the classification in the considered box, it does affect the quality of the classification in the nested box if such a nesting exists. Let us analyse this situation a bit more closely. Let $l$ and $l'$ be indices of a certain box and a nested box, respectively. The procedure to solve the problem \eqref{14} for the $l$-th box returns the set of class numbers $\mathcal{C}_l^*$. In order to solve the optimisation problem \eqref{14}  for the $l'$-th box, we need to know the class number $j_{l'}$, that is, -- based on assumption 2 -- the class related to the internal movement in the $l$-th box that opens the $l'$-th box. This class can be assigned randomly, but it can also be determined in the optimisation procedure in the $l'$-th box. In the second case, optimisation in the box $l'$ means solving the problem \eqref{14} for all classes $j_{l'}$ from the set $\mathcal{C}_l^*$ and finding the best solution. The cardinality of the solution space is now equal to the product of
$\left| \mathscr{C}_l \right|$ (see \eqref{15}) and $M_{l'}-1$. We are still dealing with a problem in which optimisation procedures are performed independently for individual boxes. The only connection occurs for nested boxes through the initial condition $jl$, which can be set randomly or selected in the optimisation procedure. It is easy to see that the situation under consideration concerns boxes of at least the second order.

\subsubsection{Optimisation problem 2}\label{sec:contextdeprecsystem:optimization:two} 

The second optimisation problem makes much more practical sense. Now, we assume that each class in $\mathcal{C}$ has exactly 2 meanings (interpretations): primary and secondary. Both meanings are exactly the same in the entire context-dependent recognition system (in all boxes and for the initial classifier). Thus, the concept of context largely loses its spectacular possibilities and is limited only to determining which of the two possible interpretations of the classification result we are dealing with in a given box. However, this is done with great benefit for user comfort and intuitiveness in the use of the bioprosthesis. At the same time, it doubles the repertoire of prosthesis movements that can be controlled, that is, $M=2C$. Since the original meaning described by formula \eqref{7} remains unchanged, our task is to give the classes $\mathcal{C} = \{1, 2, \cdots, C\}$ a secondary interpretation by assigning them an additional movement from the sequence:

\begin{equation}   \label{16}
 (m_{C+1}, m_{C+2}, \cdots , m_M). 
\end{equation}

Let us denote this particular collective interpretation (i.e., the assignment of $C$ elements) by $s(C)$ and the set of all possible secondary interpretations of classes $C$, taking into account possible constraints by $\mathcal{S}(C)$. Formally, $s(C)$ is a sequence of class numbers (a permutation of the set C):
\begin{equation}   \label{16a}
s(C)=(c_1, c_2, \cdots, c_{|C|}),
\end{equation}
the order of which corresponds to the assignment of the movements in sequence \eqref{16}.  If there were no constraints, the cardinality of $\mathcal{S}(C)$ would be factorial $C$ ($C!$).

Moreover, note that the selection of $s(C)$ from the set $\mathcal{S}(C)$ determines the sets of classes $\mathcal{C}_l$ for individual boxes, and thus affects the values of the criteria $Q_l$ that evaluate the quality of local classifications. However, now, unlike the previous optimisation problem, we cannot shape sets $\mathcal{C}_l$ separately. We can only shape them by choosing such an interpretation of $s^*(C)$ to maximise the classification quality of the entire context-dependent recognition system.
This observation leads to the following optimisation problem 2:

We are given: for each box ($l=1,2, \cdots, L$), interior movements \eqref{8} and class number $j_l$; primary class interpretations \eqref{7}; primary feature space  $\mathcal{X}$.

Goal: to find such an assignment of classes $\mathcal{C} = \{1, 2, \cdots, C\}$ to movements \eqref{16}, that is, such a sequence  $s^*(C)$ from  $\mathcal{S}(C)$ that maximises the criterion (objective function) $Q$, namely:
\begin{equation}   \label{17}
Q(s^*(C)) = \max_{s(C) \in \mathcal{S}(C)} Q(s(C)),
\end{equation} 
where the criterion (objective function) $Q$ evaluates the quality of the entire context-dependent classification system. 
We solve the problem in two steps:
\begin{enumerate}
\item Determining the set of all possible secondary meanings of individual classes $\mathcal{S}(C)$, i.e., the set of feasible solutions to optimisation problem 2. Knowledge of the primary interpretations and arrangement of the boxes, as well as the assumption that the opening and closing movements of the box are related to the same class, allows us to determine the constraints of the secondary interpretation of the classes. This is because, in the classification task specified in a given box, all classes must have different indexes (numbers). So, for each movement \eqref{16}, we can determine the permitted (non-permitted) classes that may (may not) be associated with it. The pseudocode of the iterative algorithm to determine the set of feasible solutions to the optimisation problem \eqref{17} and its cardinality is presented in Algorithm \ref{alg:Opt2}.

\item The proposed method to solve the optimisation problem \eqref{17} depends on the cardinality of the set of feasible solutions $\mathcal{S}(C)$. This cardinality does not exceed $C!$, and the smaller it gets, the more numerous the sets of non-permitted classes for movements~\eqref{16} are.   

For small problems, the exhaustive search can be used. Exhaustive search methods try to calculate all possible solutions, and then decide which one is the best. Despite these drawbacks, exhaustive search methods have a few benefits: they are simple to implement, and in the case of discrete systems, all feasible solutions are checked. 
To evaluate the solution $s(C)$, we use the scheme presented in the previous section, but this time in relation to the entire recognition system. According to this scheme, the feature selection/reduction and learning and testing procedures, as well as the model of the box classifiers, and the initial classifier are first determined, with these models and procedures then being used to calculate the value of the criterion $Q (s(C))$. 

When the optimisation problem is large and an exhaustive search is unfeasible, methods based on particular heuristics leading to approximate results can be used. In the experimental research performed, the evolutionary algorithm was used due to the well-defined components and procedures that take into account the combinatorial specificity of the optimisation problem under consideration. 
\end{enumerate}

\begin{algorithm}
{
\small
\caption{Pseudocode of the algorithm for determining the set of feasible solutions for optimisation problem 2.}\label{alg:Opt2}
\tt
\smallskip
\label{pt:tab:OP2}
\begin{tabbing}
\quad \=\quad \=\quad \=\quad \=\quad \=\quad \kill
\textbf{Input data:}\\
\>\> $\mathcal{C}_p^{(k)}=\{i_k^{(1)}, i_k^{(2)}, \cdots , i_k^{(N_k)}\}$ - set of classes \\
\>\>  permitted for $m_{C+k}$ movement,   $\mathcal{C}_p^{(k)} \subset \mathcal{C}$,   $k=1,2, \cdots, C$;\\ 
\>\> $ \mathcal{S}(1) = \mathcal{C}_p^{(1)}$ - initial values  \\
\textbf{For}  $k=2$ \textbf{to} $C$ \textbf{do}:\\
\>\> 1. Determine sets $\bar{\mathcal{C}}^{(k-1)}(i_k^{(j)})$,  $i_k^{(j)} \in \mathcal{C}_p^{(k)}$,\\
\>\> which consist of elements of the set $\bar{\mathcal{C}}^{(k-1)}$ different from $i_k^{(j)}$.  \\
\>\> The number of sets is equal to $N_k$.\\
\>\> 2. Create the Cartesian product: $\bar{\mathcal{C}}^{(k)}(i_k^{(j)})=\bar{\mathcal{C}}^{(k-1)}(i_k^{(j)}) \times i_k^{(j)}$.  \\
\>\> 3. Calculate the sum:  $\mathcal{S}(k) =  \bigcup_{i_k^{(j)}\in \mathcal{C}_p^{(k)}} \mathcal{C}^{(k)}(i_k^{(j)})$ \\
 \textbf{End for}\\
Return: $\mathcal{S}(C)$, $\left|\mathcal{S}(C)\right|$
\end{tabbing}
}
\end{algorithm}

\begin{exmp}
Let $C=5$ and $M=10$. The primary interpretation of classes is as follows: $m_1 \leftrightarrow 1, m_2 \leftrightarrow 2, m_3 \leftrightarrow 3, m_4 \leftrightarrow 4, m_5 \leftrightarrow 5$. The structure of the context classifier (box arrangement) is shown in Fig.~\ref{figs:Example1}A. A simple analysis of the classifier structure leads to the determination of a set of permitted and non-permitted classes for each movement (Fig. \ref{figs:Example1}B). Using the so-called combinatorial tree, we iteratively build all Cartesian products from unique and allowed classes starting with 2 movements ($m_6$ and $m_7$) and ending with all 5 movements $m_6 - m_{10}$. Each complete and feasible solution $c(5)$  is represented by a path in the tree from the root node to the leaf of length $C=5$ (see Fig.\ref{figs:Example1}C). It is easy to see that the number of permitted solutions (or secondary meaning of classes) is equal to 12. Such a cardinality of a set of solutions justifies the use of the exhaustive search method.  

The maximum possible cardinality of the set $\mathcal{S}(5)$ is 5!=120. 
A radical reduction in the number of elements results from the constraints defined in the sets of non-permitted classes. With a specific arrangement of non-permitted classes, it may happen that the set $\mathcal{S}(C)$ will be an empty set. In such a situation, we need to rearrange the structure of boxes in the context-dependent classifier.

\begin{figure}[hbt]
	\centering
		\includegraphics[width=0.5\textwidth]{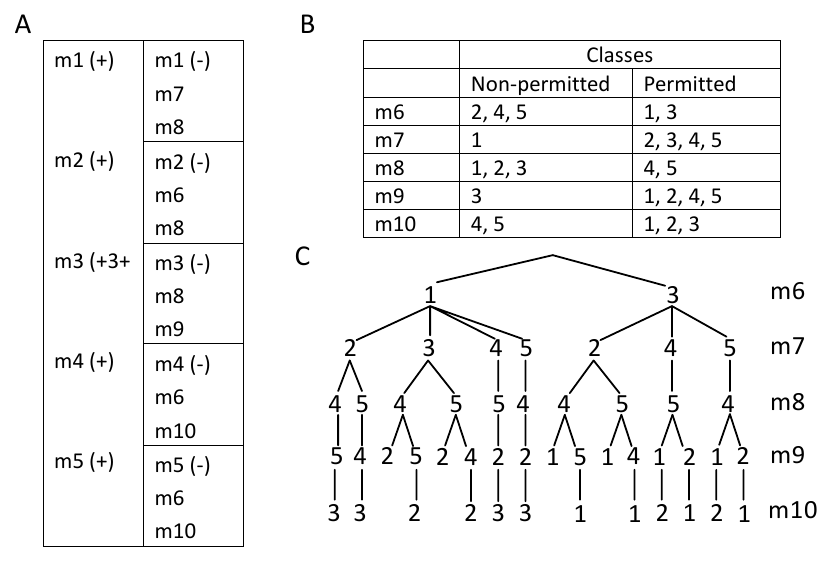}
	\caption{Example 1:  (A) structure of the context-dependent classification system (box arrangement); (B) permitted and non-permitted class numbers for the secondary interpretation; (C) the solution tree.  
}
	\label{figs:Example1}
\end{figure}

\end{exmp} 

\begin{exmp}
Data ($C$, $M$ and primary interpretation) are the same as in Example 1. The example illustrates a situation where among the classes non-permitted for individual movements $m_6 - m_{10}$ there are \textit{a priori} unknown class numbers. They can only be determined in the subsequent steps of the procedure, which leads to variants of the solution paths.

\begin{figure}[hbt]
	\centering
		\includegraphics[width=0.5\textwidth]{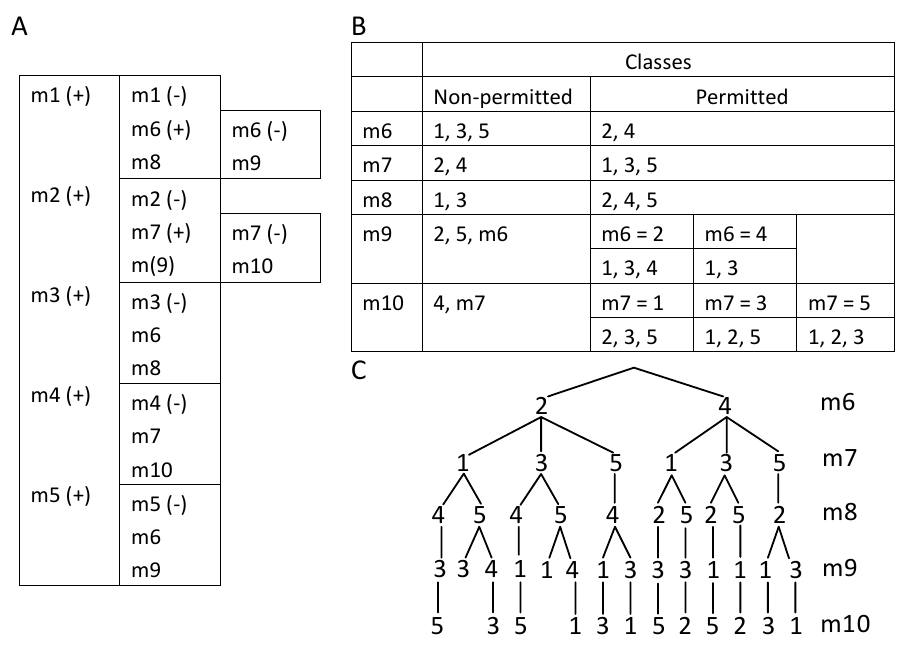}
	\caption{Example 2:  (A) the structure of the context-dependent classification system (box arrangement); (B) permitted and non-permitted class numbers for the secondary interpretation; (C) the solution tree. 
}
	\label{figs:Example2}
\end{figure}
\end{exmp}

\subsection{Evolutionary algorithm for the solution of optimisation problem 2}

In the conducted experimental evaluation, the evolutionary algorithm was proceeded as follows:
\begin{enumerate}
\item \textbf{Representation.} Since the solution to the problem \eqref{17} has the form of the sequence \eqref{16a}, the most natural representation of a chromosome (individual) is a permutation of a fixed set of classes $\mathcal{C}$, that is, a set of natural numbers $\{1, 2, \cdots , C\}$. 

\item \textbf{Evaluation.} The fitness function is exactly the same as the objective function $Q$ in the optimisation problem \eqref{17}. 
\item \textbf{Initialisation.} The population of a given size $P_{size}$ is randomly selected with a uniform distribution. The size of the population is the same in all generations. If after $T$ generations the solutions were evaluated without improvement, then replace the solutions in the next generation with $P_{size}-1$ random solutions and the best solution for the $T$-th population.

\item \textbf{Parent Selection.} Tournament procedure was applied as the parent selection method. Tournament selection is conceptually simple, fast to implement, and does not require any global knowledge of the population. 

\item \textbf{Variation Operators.} 
\begin{enumerate}
\item \textbf{Recombination.} There are two types of optimisation problem that are represented by permutations: problems in which the \emph{order} or \emph{adjacency} of events is important. The problem \eqref{17} belongs to the first group, therefore, the order crossover methods OX1 and OX2 were selected.

\item \textbf{Mutation.} Swap mutation, Insert mutation, Scramble mutation, Inversion mutation were applied.
\end{enumerate}

\item \textbf{Repair Procedure.}  For permutation
code, ordinary crossover and mutation operators might generate illegal chromosomes. Repair is a way to handle infeasible individuals. In the proposed method, infeasible offspring are replaced by the nearest feasible solutions. Kendall's tau distance (the number of discordant pairs) \cite{Cicirello2019} was used as a measure of the distance between two permutations \eqref{16a}.

\item \textbf{Survivor selection (Replacement).} The fitness-based replacement with elitism was applied.

\item \textbf{Termination Condition (Stop procedure).} The evolution process was terminated after a fixed number of generations.
\end{enumerate}

\section{Experimental Setup}\label{sec:experiments}\label{sec:experiments:setup}


The experiments are conducted in order to answer the following research questions: \textbf{RQ1} Does the context-dependent classifier achieve a better or similar classification quality to the context-free classifier? \textbf{RQ2} Does the box-structure optimisation (selecting proper movement-to-class binding) allow the context-dependent method to perform better when compared to a random (among the feasible solutions) box-structure choice?

Answering the research questions implies comparing the three following methods: 

\begin{enumerate}

    \item \textbf{Plain}. This is a context-free classifier that just recognises all classes from set $\mathcal{C}$. This is a baseline method to which the other methods are compared. 

    \item \textbf{RCtx}. This is a context-dependent classifier that uses a randomly chosen movement-to-class binding. The set of feasible solutions is determined using Algorithm~\ref{alg:Opt2}. The context-dependent classifier is, in fact, a classifier ensemble. The members of the ensemble (called base classifiers) are used to predict a set of box-dependent classes $\mathcal{C}_l$. When a prediction is made, the context is switched according to the box structure implemented by the context-dependent classifier. Examples of context-dependent structures are presented in section~\ref{sec:contextdeprecsystem:optimization:two}. The box-specific attribute set $\bar{ \mathcal{X}_l }$  is selected using a filter-based feature selection procedure that uses mutual information to assess attributes~\cite{Zebari2020}. When attributes are assessed, we choose 50\% of the best-scored attributes to train a box-specific classifier.

    \item \textbf{OCtx}. This is a context-dependent classifier that uses the best movements-to-classes binding among the solutions generated using the Algorithm~\ref{alg:Opt2}. The quality of the context-dependent structure is determined by building the context-dependent classifier and then evaluating it in terms of the sequence coverage (\textbf{SqCov}) criterion (defined later in this section). Box-specific classifiers are trained in the same way as described for \textbf{RCtx}. Training and testing sets are created using three-fold cross-validation. The quality criterion and the detailed description of the evaluation procedure are provided later in this chapter. 

\end{enumerate}

During the experimental study, we consider the following classifiers in order to build context-dependent ensembles and context-free classifiers:  Random Forest classifier (\textbf{RF}) with the committee size set to 20; Naive Bayes classifier (\textbf{NB}) employing Gaussian distribution for modelling the class-conditional distributions; Nearest Neighbour classifier (\textbf{NN}); SVM Classifier with linear kernel (no kernel), (\textbf{SVL}) with one-vs-one multiclass decomposition; SVM Classifier with RBF kernel (\textbf{SVR}) with one-vs-one multiclass decomposition.
      
We use the classifiers implemented in the scikit-learn package~\cite{scikit-learn}. If not stated otherwise, the classifier parameters are set to their default values.

The ensembles are homogeneous. It means that for the ensemble, only one classification algorithm is used as the base classifier. Multiple models of base classifiers are considered when checking whether the choice of the box-specific classification procedure changes the outcome of the experiment.

\begin{figure}[tb]
    \centering
        \includegraphics[width=0.8\textwidth]{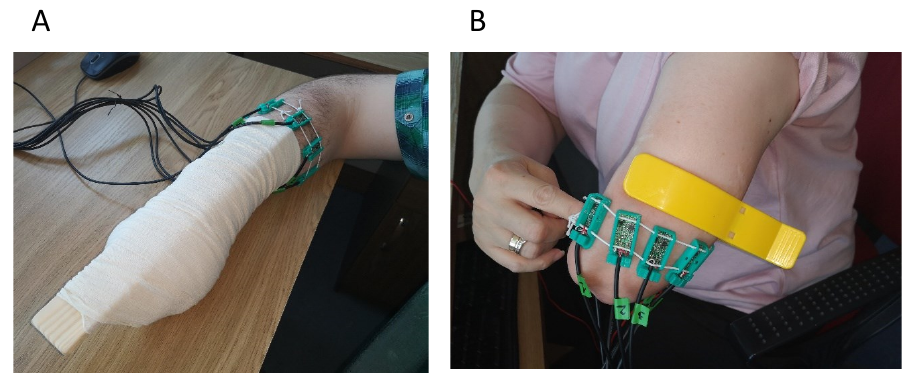}
    \caption{Illustration of  recording sEMG signals for  signalset 1 (A) and signalset 2 (B)}
    \label{figs:Photos}
\end{figure}

\begin{figure}[hbt]
	\centering
		\includegraphics[width=0.7\textwidth,height=0.3\textheight]{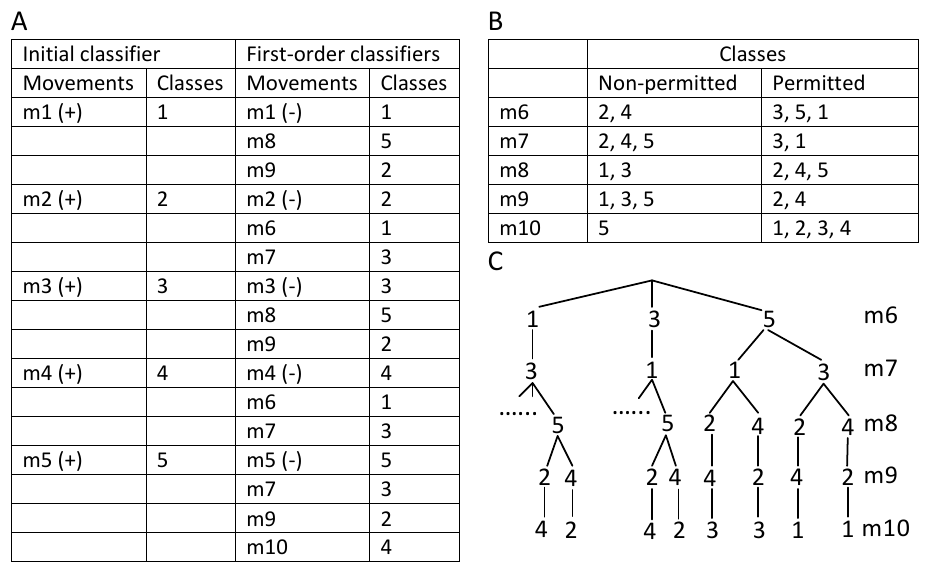}
	\caption{Signalset 2:  (A) the structure of the context-dependent classification system (box arrangement); (B) permitted and non-permitted class numbers for the secondary interpretation; (C) the solution tree.}
	\label{figs:Basia}
\end{figure}

\begin{figure}[hbt]
	\centering
		\includegraphics[width=0.7\textwidth, height=0.4\textheight, keepaspectratio]{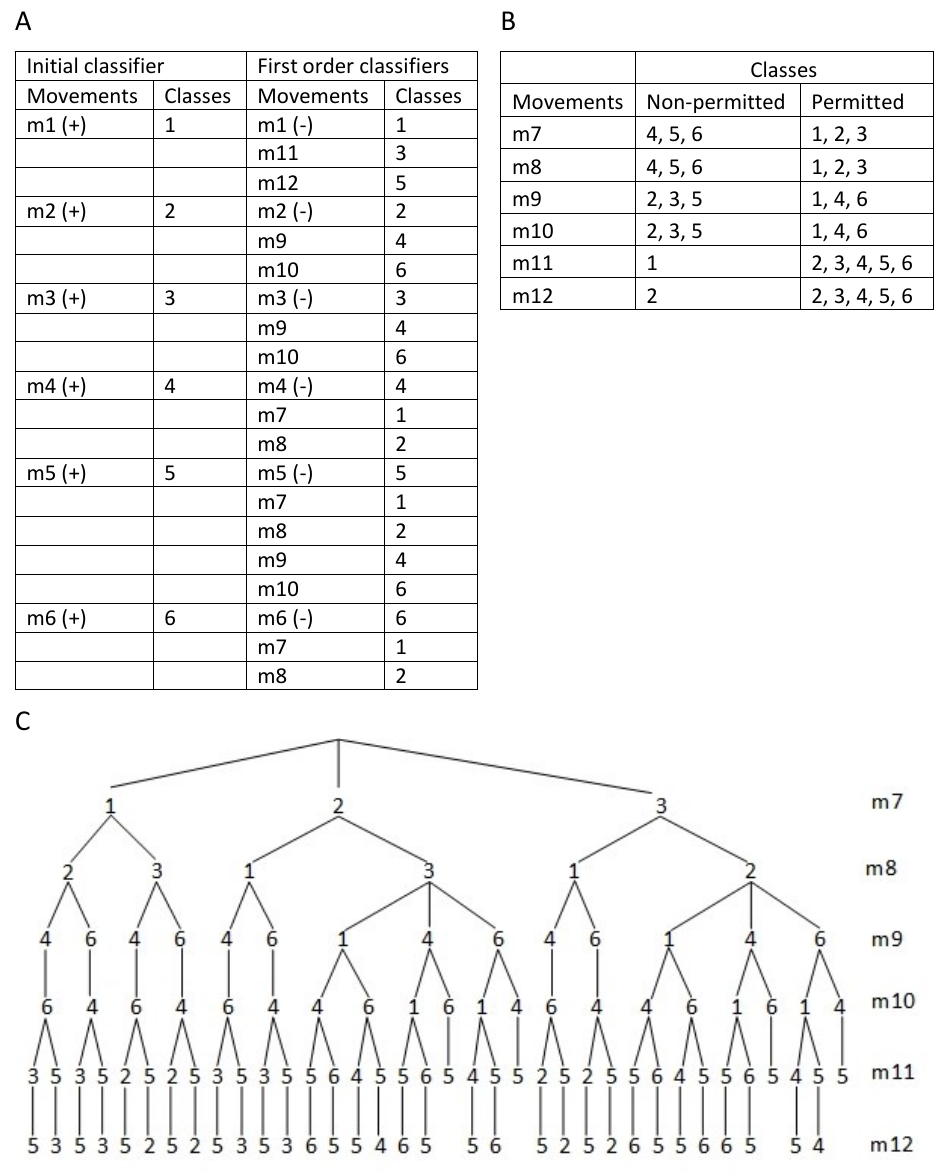}
	\caption{Signalset 3:  (A) the structure of the context-dependent classification system (box arrangement); (B) permitted and non-permitted class numbers for the secondary interpretation; (C) the solution tree. }
	\label{figs:SixClasses}
\end{figure}

The experiments were conducted using the following signalsets:

\begin{itemize}
    \item \textbf{Signalset 1}. A male able-body  subject, aged 73, right-handed. When creating the set of signals, a simulation of amputation was modelled by completely immobilising the hand during the recording of signals, which obliges the subject to generate EMG signals based on his own image of the movement performed, and not on the basis of actual movements performed by the hand~\cite{Schone2023}.
The sEMG biosignals were registered using a special designated measuring circuit with 8 sensors evenly spaced around the forearm and a sampling frequency of 1000 Hz. The signalset used in the experiments consisted of eight classes (according to the subject's imagination): (1) wrist flexion, (2) wrist extension, (3) ulnar deviation, (4) radial deviation, (5) index and middle fingers flexion, (6) index and middle fingers extension, (7) ring and little fingers flexion, (8) ring and little fingers extension. One hundred 8-channel sEMG signals per class were recorded. Each measurement lasted 1000 ms and was preceded with a 3 s break.
For the recognition problem with 8 classes, 16 prosthesis movements were selected and the structure of the context-dependent recognition system presented in Fig.~\ref{figs:UR} was proposed. The sensor placement and the method of hand immobilization are shown in Fig.~\ref{figs:Photos}A. Subject gave informed consent prior to participation.

\item \textbf{Signalset 2}. A female amputee, 46 years old, right-handed, with traumatic deficiency below the elbow on the left side.  
The measuring circuit, the number of sensors (channels) and a sampling frequency as in signalset 1 were used. The set of signals used in the experiments consisted of five classes (subject's imagination of the movement of the phantom hand): (1) precision grip, (2) lateral grip, (3) hook grip, (4) spherical grip, and (5) cylindrical grip. Sixty 8-channel sEMG signal samples per class were recorded. Each measurement lasted 1000 ms and was preceded by a 5 s break. 
Fig.~\ref{figs:Basia} shows the proposed structure of the context-dependent  classifier for signalset 2, i.e. for a task with 5 classes and 10 prosthesis movements. For ease, the primary interpretation of classes is identical to phantom movements. As a secondary interpretation of the classes, the following 5 prosthesis movements have been selected:  supination (m6), pronation (m7), wrist flexion (m8) wrist extension (m9) and radial deviation (m10). The sensor placement is shown in Fig.~\ref{figs:Photos}B. Subject gave informed consent prior to participation.

\item \textbf{Signalset 3}. Signalset 3 comes from the web repository \footnote{\url{https://www.rami-khushaba.com/}}. sEMG signals were collected from nine transradial amputees, 7 male, 2 female, 19 - 35 years old. The demographic information details for each amputee are presented in \cite{AlTimemy2016}.
A common signal acquisition protocol was used for all the participants: 6 classes
(imagination of movement of fantom hand) - (1) thumb flexion, (2) index flexion, (3) fine pinch, (4) tripod grip, (5) hook grip, (6) spherical grip; 8 - 12 sEMG sensors (depending on the diameter of the forearm stump); sampling frequency 2 kHz; for each of the 6 classes, sEMG was recorded for 3 force levels: low force, moderate force, high force; for each class and for each force level, 5 - 8 trials were recorded, including the holding phase, and lasting 8 - 12 s.  
To unify signalset 3, we used sEMG signals from 8 channels and which were associated with a low force level. In separating individual objects for particular classes, we used a non-overlapped segmentation scheme with a segment length of 500 ms. This resulted in 70 to 310 objects (eight-channel sEMG signal records) for each class. Fig.~\ref{figs:SixClasses} shows a common context-dependent classifier scheme with 6 classes and 12 movements applied to all 9 amputees. The selected movements are as follows: mouse grip (m1), hook grip (m2), platform grip (m3), key grip (m4), precision grip (m5), cylindrical grip (m6), supination (m7), pronation (m8), wrist flexion (m9), wrist extension (m10), ring finger flexion (m11), index finger flexion (m12).

\end{itemize}

For each signalset, a separate context-dependent classifier structure was proposed. To avoid combining results related with different context-dependent structures, the experiments for each set of signals are analysed separately. Signalsets 1 and 2 are related to a single subject, and therefore the results are presented for a single subject only. Signalset 3 contains data related to nine subjects. In this case, results for a single subject are also presented.  Additionally, the results for all subjects are analysed as a whole. To do so, the average rank~\cite{garcia2008extension} approach is used to aggregate subject-specific results. For all signalsets, we also presented the results aggregated (using the average ranks approach) over all the base-classifier types used to build the evaluated models. 

\begin{figure}[!htb]
	\centering
		\includegraphics[width=0.6\textwidth, height=0.45\textheight, keepaspectratio]{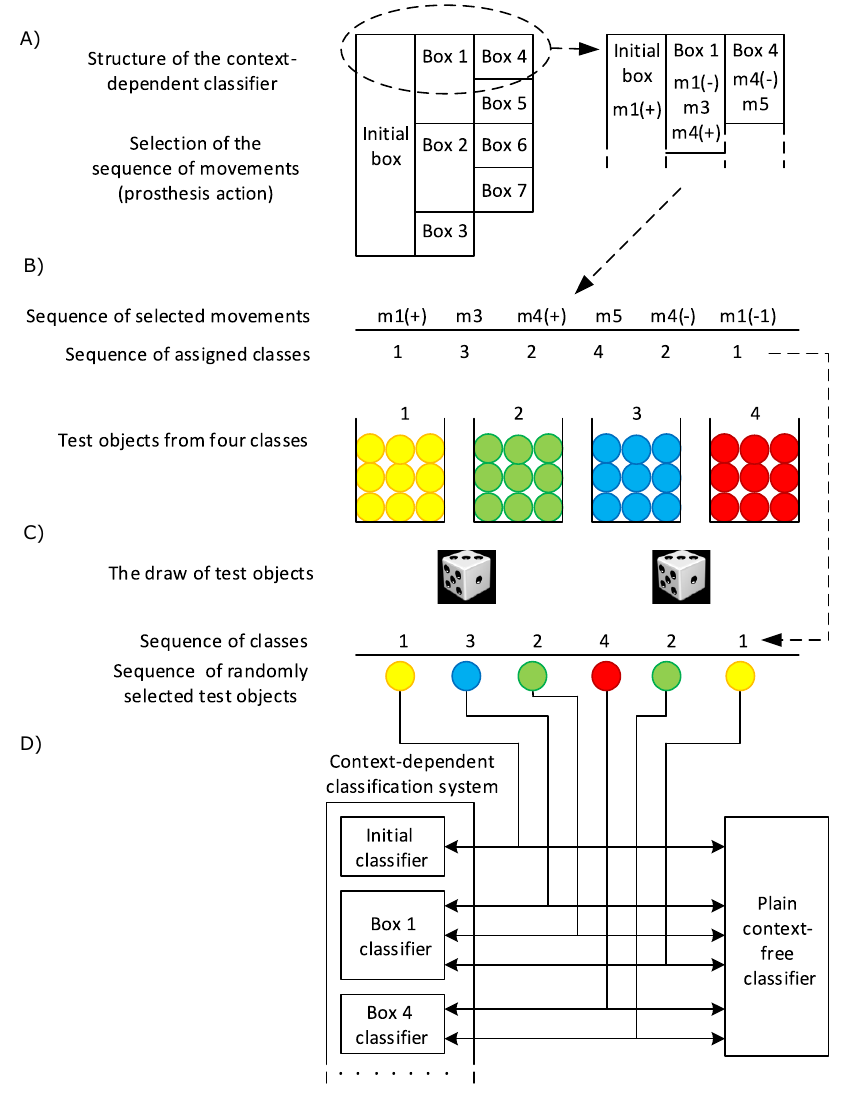}
	\caption{Scheme of testing the classifiers}
	\label{figs:Experiment}
\end{figure}

The training procedure of the context-dependent classifier is described in section~\ref{sec:contextdeprecsystem:model}. Due to the context-dependent nature of the presented methods, the testing procedure is a bit different from the ordinary classifier-testing procedure. Since the result of inside-box classification can change context (the change of context is as simple as moving to the different boxes inside the ensemble context-dependent structure), the order of test samples is very important. This test sample order decides which boxes will be activated inside the ensemble structure. To ensure that all boxes are activated during the testing phase, we proposed the testing procedure illustrated in Fig.~\ref{figs:Experiment}. The context structure of the ensemble classifier is known for each signalset. By knowing this structure, we may generate movement sequences that start in the $0\mathrm{-th}$ order box (initial box), pass through intermediate boxes, and end with a leaf box (box without other boxes nested in it). This is shown in Fig.~\ref{figs:Experiment}A. The number of the movement sequences will be denoted as $G$. This number depends on the box-structure of the ensemble. Such a sequence activates (makes a box-specific classifier make a prediction for a testing object) all the boxes along the path. To guarantee that all inside-box classifiers in the ensemble are activated, we must generate sequences that cover all possible paths from a zero-order box to the leaves of the tree-like structure. When the ensemble is trained, the mapping between movements and classes is also known. These movement sequences can be converted into sequences containing class labels (See Fig.~\ref{figs:Experiment}B). When the sequence of classes is known, we can randomly (with replacement) choose class-specific objects from the testing set (Fig.~\ref{figs:Experiment}C). To provide a better approximation for sequence-specific classification quality, for the given class sequence, the random choice of class-specific objects for the sequence is repeated $R=20$ times. This gives us $R$ object sequences for each movement sequence. Finally, the sequences of objects are passed to the classifiers, and the values of the quality criteria are calculated (Fig.~\ref{figs:Experiment}D). To make the sequence classification tasks independent, the state of the context-dependent ensemble is set to the initial state after the sequence of objects is fed to the ensemble. The initial state is the state when the zero-order box-specific classifier awaits objects. The context-free classifier (\textbf{Plain}) is evaluated in the same way, but the sequence of the object is generated using the movement-to-class binding taken from \textbf{RCtx}. This is to make the context-free and context-dependent classifiers comparable.

To create the training and testing sets, the stratified, ten-fold cross-validation procedure is applied. The discrete wavelet transformation technique was used to create feature vectors from a raw sEMG signal. We used the \textit{db6} wavelet and three levels of decomposition. The following functions were calculated for the transformation coeffcients \cite{MendesJunior2020}: MAV -- Mean absolute value, SSC -- Slope sign change, Coefficients of the 3rd order, linear, autoregressive model~\cite{Suplino2019}.

\phantomsection\label{phantom:seqcov}
To evaluate the classification outcome for the sequences of objects, the subject/user perspective must be taken into account. Traditional classification quality measures such as accuracy or $F_1$ are not suitable here. This is because traditional measures count hits and misses for a single object, and then calculate the summary value. In our case, we must concentrate on the entire testing sequence. From the subject's perspective, the prosthesis works correctly if the entire sequence of movements is performed accurately. In other words, as a result of user input, the prosthesis has moved to the desired state/pose. This leads to the quality measure similar to the zero-one measure for multi-label learning~\cite{Luaces2012}. Basically, if the classification is correct for each object in the sequence, then the value of the zero-one metric is one. If the classification is incorrect for even a single element of the sequence, the zero-one metric value is zero. In the paper, we denote this quality measure by \textbf{ZO}. 

However, this measure is a bit coarse. It cannot distinguish between the sequence of objects that is almost perfectly recognised (i.e., all but one object are classified correctly) as well as in the sequence in which all objects are misclassified. Consequently, we also need a criterion that is able to measure sequence quality using more values than zero or one. However, simply counting misses and hits (accuracy) is not suitable here. When the subject uses the prosthesis, the first misclassification forces him/her to stop generating muscle contractions and retract the last change in the state of the prosthesis. Consequently, the moment of the first misclassification seems to be the most important one. The moment of the first misclassification is also strongly related to the cumulative probability of making the misclassification at a given position along the sequence. Taking this into account, we propose the \textbf{SqCov} quality measure. It counts the number of correctly classified sequence elements until the first misclassification occurs. Then, the number of correctly classified objects is divided by the length of the entire sequence. Consequently, the value of the criterion is in the interval $[0;1]$. 

To provide a formal description of the proposed quality measures, the following notation is introduced. Let $K=GR$ be the total number of sequences checked during the testing phase. Consequently, the number of objects in the $k\mathrm{-th}$ sequence (length of the sequence) is denoted by $L_k, \; k \in \{ 1, 2, \cdots, K\}$. The indexes of objects within the sequence are in $\{ \{ 1, 2, \cdots, K\} \}$. The zero-one (\textbf{ZO}) criterion may be calculated as follows:
\begin{align}\label{eq:zero-one}
\Delta_{k}^{(z)} &= 
\begin{cases}
1 & \text{if no classification errors are made} \\
0 & \text{otherwise}
\end{cases}
\\
\mathrm{ZO} &= \frac{1}{K} \sum_{k=1}^{K}\Delta_{k}^{(z)}.
\end{align}

The position of the first misclassified object in the $k\mathrm{-th}$ sequence will be denoted as $l_k,\; k \in \{ 1, 2, \cdots, K\}$. Given that, the \textbf{SqCov} criterion is defined using the following formula:
\begin{align}\label{eq:seq-cov}
\Delta_{k}^{(s)} &= 
\begin{cases}
1 & \text{if no classification errors are made} \\
\frac{l_k - 1 }{L_k} & \text{if } l_k \leq L_k
\end{cases}
\\
\mathrm{SqCov} &= \frac{1}{K} \sum_{k=1}^{K}\Delta_{k}^{(s)}.
\end{align}

Following the recommendations of~\cite{garcia2008extension}, the statistical significance of the results obtained was evaluated using the pairwise Wilcoxon signed rank test. Since multiple comparisons were made, family-wise errors (FWER) should be controlled. To do so, Holm's procedure of controlling FWER was employed. For all the tests, the significance level was set to $\alpha=0.05$. We employed statistical tests and correction procedures employed in SciPy~\cite{Virtanen2020}.

\section{Results}\label{sec:resultsanddisc}

The experimental results are presented in Tables~\ref{table:Set1_res} -- \ref{table:Set3_avg_ranks_c}. There are two types of results presented. Tables~\ref{table:Set1_res},~\ref{table:Set2_res} and \ref{table:Set3_patient1_res} -- \ref{table:Set3_patient9_res} present the average criterion value and standard deviation calculated for all test folds. The tables are divided into two sections with regards to quality criteria. In each section, there are three columns. Each of the columns is related to one of the investigated methods, each row is related to one of the base classifiers, and each cell in the table contains two types of information. The first is the average and the standard deviation. The second one, written in subscripts, contains the method numbers for which the indicated method is significantly (according to the statistical test performed) better. There can also be a hyphen written in the subscript. It means that there is no significant difference from other methods. The method numbers are presented in section~\ref{sec:experiments:setup}.  Tables \ref{table:Set1_res_c}, \ref{table:Set2_res_c}, \ref{table:Set3_avg_ranks} and \ref{table:Set3_avg_ranks_c} follow the same format, but instead of mean values and standard deviations, they present the averaged ranks for all subjects. For a single subject, the ranks are assigned as follows. The method that achieves the highest average of the criterion value gets rank 3 (the number of methods to compare), the second one 2, and the last one rank 1. The interpretation of the subscript entries is the same.


Let us begin with the analysis of results for signalset 1. The results presented in Table~\ref{table:Set1_res_c} clearly show, for all base classifiers, that \textbf{OCtx} achieves the highest values of both quality criteria. \textbf{RCtx} is second, and \textbf{Plain} achieves the lowest values of quality criteria. However, after performing the statistical test on the aggregated data, there are no significant differences between the models. This is probably due to the small number of base classifiers. If we analyse the base-classifier-specific results, some significant differences may be found. In most cases, \textbf{RCtx} and \textbf{OCtx} are significantly better than \textbf{Plain}. For other cases, there are no significant differences between the investigated methods. This means that context-dependent ensembles perform similarly or better than the context-free one. Consequently, switching to the context-dependent model does not harm the classification quality of the movement intentions. Additionally, the context-dependent ensembles presented in this paper allow us to recognise twice as many movement intentions as the context-free one. This is because, in this study, we limit the number of class meanings to two (see Section~\ref{sec:contextdeprecsystem:optimization:two}). However, if the user/subject accepts the binding of more movement classes into a single muscle-contraction pattern (class), the effective number of recognised movement intentions may be even greater. Consequently, the dexterity of the prosthesis can be increased without any reduction in classification quality. For \textbf{SVL}, \textbf{NB} and \textbf{NN}, \textbf{OCtx} is significantly better than \textbf{RCtx}. This fact, together with the average ranks presented, means that the optimisation of the movement-to-class binding is an important step in building the context-dependent ensemble, as it may improve the overall classification quality. However, even a random choice of this binding may give satisfactory results.

\begin{table}[htb]
\renewcommand{\arraystretch}{0.6}
\centering
\footnotesize
\caption{Classification quality for signalset 1.\label{table:Set1_res}}
\begin{tabular}{llll}
 & \multicolumn{3}{c}{SqCov} \\
 & Plain & OCtx & RCtx \\
\cmidrule(lr){2-4}\\
SVL & .962 $\pm$ .018 & .983 $\pm$ .009 & .969 $\pm$ .019 \\
         & -- & {\scriptsize 1,3} & -- \\
SVR & .968 $\pm$ .022 & .981 $\pm$ .014 & .977 $\pm$ .014 \\
 & -- & {\scriptsize 1} & {\scriptsize 1} \\
NB & .840 $\pm$ .036 & .908 $\pm$ .024 & .887 $\pm$ .026 \\
 & -- & {\scriptsize 1,3} & {\scriptsize 1} \\
NN & .751 $\pm$ .059 & .851 $\pm$ .028 & .840 $\pm$ .035 \\
 & -- & {\scriptsize 1} & {\scriptsize 1} \\
RF & .915 $\pm$ .047 & .941 $\pm$ .026 & .938 $\pm$ .026 \\
 & -- & {\scriptsize 1} & -- \\
\end{tabular}%
\begin{tabular}{lll}
 \multicolumn{3}{c}{ZO} \\
  Plain & OCtx & RCtx \\
\cmidrule(lr){1-3}\\
 .942 $\pm$ .028 & .980 $\pm$ .012 & .953 $\pm$ .028 \\
 -- & {\scriptsize 1,3} & -- \\
 .952 $\pm$ .034 & .977 $\pm$ .022 & .967 $\pm$ .021 \\
 -- & -- & -- \\
 .772 $\pm$ .046 & .906 $\pm$ .024 & .852 $\pm$ .039 \\
 -- & {\scriptsize 1,3} & {\scriptsize 1} \\
 .639 $\pm$ .080 & .816 $\pm$ .033 & .789 $\pm$ .049 \\
 -- & {\scriptsize 1,3} & {\scriptsize 1} \\
 .873 $\pm$ .071 & .928 $\pm$ .033 & .921 $\pm$ .034 \\
 -- & {\scriptsize 1} & {\scriptsize 1} \\
\end{tabular}

\end{table}

\begin{table}[htb]
\renewcommand{\arraystretch}{0.6}
\centering
\footnotesize
\caption{Classification quality for signalset 1. Average ranks over base classifiers\label{table:Set1_res_c}}

\begin{tabular}{llll}
 & \multicolumn{3}{c}{SqCov} \\
 & Plain & OCtx & RCtx \\
\cmidrule(lr){2-4}\\
Avg Rnk & 1.000 & 3.000 & 2.000 \\
 & -- & -- & -- \\
\end{tabular}%
\begin{tabular}{lll}
 \multicolumn{3}{c}{ZO} \\
  Plain & OCtx & RCtx \\
\cmidrule(lr){1-3}\\
1.000 & 3.000 & 2.000 \\
 -- & -- & -- \\
\end{tabular}

\end{table}


The results for signalset 2 are quite similar to those from signalset 1, but slightly less optimistic. The fold-averaged values of quality criteria for all base classifiers are significantly lower than the analogous ones from signalset 1. This is probably because signalset 1 was taken from an able-bodied person, whereas signalset 2 was taken from a person with an amputation. The other cause may be a significant difference in sample size. For signalset 2, we have only 60 objects per class compared to 100 objects per class in signalset 1. A smaller sample number with a lower signal-to-noise ratio may have led to the classifiers overfitting. But still, in the best case of the \textbf{RF} classifier, the proposed classification system offers (in terms of ZO measure) classification quality above 68\%. However, even if the overall classification quality is lower, we may still benefit from using context-dependent ensembles in this scenario. In general, context-dependent methods are comparable to the context-free one. For some base classifiers, they are significantly better, and since they offer a greater number of movements to use, they are still preferable over the \textbf{Plain} method.

\begin{table}[htb]
\renewcommand{\arraystretch}{0.6}
\centering
\footnotesize
\caption{Classification quality for signalset 2.\label{table:Set2_res}}

\begin{tabular}{llll}
  & \multicolumn{3}{c}{SqCov} \\
 & Plain & OCtx & RCtx \\
\cmidrule(lr){2-4}\\

SVL & .561 $\pm$ .107 & .590 $\pm$ .076 & .610 $\pm$ .073 \\
 & -- & -- & -- \\
SVR & .699 $\pm$ .126 & .707 $\pm$ .124 & .700 $\pm$ .092 \\
 & -- & -- & -- \\
NB & .477 $\pm$ .109 & .533 $\pm$ .117 & .523 $\pm$ .103 \\
 & -- & {\scriptsize 1} & {\scriptsize 1} \\
NN & .486 $\pm$ .091 & .546 $\pm$ .104 & .559 $\pm$ .121 \\
 & -- & {\scriptsize 1} & {\scriptsize 1} \\
RF & .769 $\pm$ .053 & .753 $\pm$ .060 & .753 $\pm$ .090 \\
 & -- & -- & -- \\
\end{tabular}%
\begin{tabular}{lll}
 \multicolumn{3}{c}{ZO} \\
 Plain & OCtx & RCtx \\
\cmidrule(lr){1-3}\\
 .428 $\pm$ .125 & .470 $\pm$ .101 & .500 $\pm$ .083 \\
 -- & -- & -- \\
 .589 $\pm$ .164 & .619 $\pm$ .152 & .606 $\pm$ .115 \\
  -- & -- & -- \\
 .327 $\pm$ .119 & .417 $\pm$ .139 & .403 $\pm$ .111 \\
 -- & {\scriptsize 1} & {\scriptsize 1} \\
 .327 $\pm$ .102 & .424 $\pm$ .103 & .434 $\pm$ .128 \\
 -- & {\scriptsize 1} & {\scriptsize 1} \\
 .677 $\pm$ .066 & .683 $\pm$ .069 & .677 $\pm$ .107 \\
 -- & -- & -- \\
\end{tabular}
ļ
\end{table}

\begin{table}[htb]
\renewcommand{\arraystretch}{0.6}
\centering
\footnotesize
\caption{Classification quality for signalset 2. Average ranks over base classifiers\label{table:Set2_res_c}}

\begin{tabular}{llll}
 & \multicolumn{3}{c}{SqCov} \\
 & Plain & OCtx & RCtx \\
\cmidrule(lr){2-4}\\
Avg Rnk & 1.400 & 2.400 & 2.200 \\
& -- & -- & -- \\
\end{tabular}%
\begin{tabular}{llll}
 \multicolumn{3}{c}{ZO} \\
 Plain & OCtx & RCtx \\
\cmidrule(lr){1-3}\\
 1.200 & 2.600 & 2.200 \\
 -- & -- & -- \\
\end{tabular}

\end{table}

The direct comparison of signalset 3 with signalsets 1 and 2 is less accurate due to the different measurement circuits used. However, some valuable conclusions can also be drawn. As we can see, the classification quality is significantly higher when compared to the quality achieved by the methods for signalset 2. For both quality criteria and all base classifiers, the values are above 0.75. The reason for these differences may be biological (subject dependent), or may lay in the different sample sizes (for signalset 3 the number of instances for a class is higher).  For each subject, the differences between \textbf{Plain}, \textbf{RCtx}, and \textbf{OCtx} follow a similar pattern as in signalsets 1 and 2. That is, context-dependent methods are comparable to or even better than the context-free one. For some subjects and some base classifiers, \textbf{OCtx} is significantly better than \textbf{RCtx}. When analysing the aggregated results presented in Table~\ref{table:Set3_avg_ranks}, a clearer picture emerges. It can be seen that in terms of average ranks, \textbf{OCtx} performs better than \textbf{RCtx} and \textbf{Plain}. \textbf{RCtx} also achieves higher ranks than \textbf{Plain}. In most cases, these differences are significant. When we calculate the average ranks for all subjects and base classifiers, it can be seen that the context-dependent methods are significantly better than the context-free one. Furthermore, \textbf{ OCtx} is significantly better than \textbf{RCtx}. The overall results for the nine-subject cohort and five base classifiers suggest that context-dependent models should be preferred over context-free methods. 

\begin{table}[htb]
\renewcommand{\arraystretch}{0.6}
\centering
\footnotesize
\caption{Classification quality for dataset 3, subject 1.\label{table:Set3_patient1_res}}
\begin{tabular}{llll}
 & \multicolumn{3}{c}{SqCov} \\
 & Plain & OCtx & RCtx \\
 \cmidrule(lr){2-4}\\
SVL & .997 $\pm$ .008 & .997 $\pm$ .008 & .996 $\pm$ .009 \\
 & -- & -- & -- \\
SVR & .983 $\pm$ .018 & .985 $\pm$ .017 & .982 $\pm$ .019 \\
 & -- & -- & -- \\
NB & .934 $\pm$ .039 & .942 $\pm$ .041 & .936 $\pm$ .040 \\
 & -- & -- & -- \\
NN & .932 $\pm$ .042 & .952 $\pm$ .030 & .939 $\pm$ .029 \\
 & -- & {\scriptsize 1,3} & -- \\
RF & .926 $\pm$ .061 & .952 $\pm$ .029 & .952 $\pm$ .045 \\
 & -- & -- & -- \\
\end{tabular}%
\begin{tabular}{lll}
 \multicolumn{3}{c}{ZO} \\
  Plain & OCtx & RCtx \\
  \cmidrule(lr){1-3}\\
 .996 $\pm$ .012 & .995 $\pm$ .011 & .993 $\pm$ .013 \\
 -- & -- & -- \\
 .975 $\pm$ .027 & .980 $\pm$ .023 & .975 $\pm$ .026 \\
 -- & -- & -- \\
 .903 $\pm$ .058 & .923 $\pm$ .052 & .909 $\pm$ .059 \\
 -- & -- & -- \\
 .900 $\pm$ .061 & .933 $\pm$ .045 & .919 $\pm$ .039 \\
 -- & {\scriptsize 1} & -- \\
 .891 $\pm$ .086 & .937 $\pm$ .032 & .933 $\pm$ .056 \\
 -- & -- & -- \\
\end{tabular}
\end{table}

\begin{table}[htb]
\renewcommand{\arraystretch}{0.6}
\centering
\footnotesize
\caption{Classification quality for dataset 3, subject 2.\label{table:Set3_patient2_res}}
\begin{tabular}{llll}
 & \multicolumn{3}{c}{SqCov} \\
 & Plain & OCtx & RCtx \\
 \cmidrule(lr){2-4}\\
SVL & .993 $\pm$ .015 & .994 $\pm$ .012 & .994 $\pm$ .012 \\
 & -- & -- & -- \\
SVR & .988 $\pm$ .014 & .990 $\pm$ .012 & .991 $\pm$ .011 \\
 & -- & -- & -- \\
NB & .967 $\pm$ .025 & .974 $\pm$ .021 & .972 $\pm$ .019 \\
 & -- & -- & -- \\
NN & .991 $\pm$ .013 & .993 $\pm$ .011 & .991 $\pm$ .013 \\
 & -- & -- & -- \\
RF & .960 $\pm$ .027 & .972 $\pm$ .024 & .971 $\pm$ .020 \\
 & -- & -- & -- \\
\end{tabular}%
\begin{tabular}{lll}
 \multicolumn{3}{c}{ZO} \\
  Plain & OCtx & RCtx \\
  \cmidrule(lr){1-3}\\
 .990 $\pm$ .021 & .992 $\pm$ .016 & .993 $\pm$ .014 \\
 -- & -- & -- \\
 .983 $\pm$ .021 & .986 $\pm$ .017 & .988 $\pm$ .016 \\
  -- & -- & -- \\
 .952 $\pm$ .038 & .966 $\pm$ .028 & .962 $\pm$ .024 \\
  -- & -- & -- \\
.987 $\pm$ .020 & .991 $\pm$ .014 & .988 $\pm$ .018 \\
 -- & -- & -- \\
 .940 $\pm$ .039 & .960 $\pm$ .032 & .962 $\pm$ .027 \\
  -- & -- & -- \\
\end{tabular}
\end{table}

\begin{table}[htb]
\renewcommand{\arraystretch}{0.6}
\centering
\footnotesize
\caption{Classification quality for dataset 3, subject 3.\label{table:Set3_patient3_res}}
\begin{tabular}{llll}
 & \multicolumn{3}{c}{SqCov} \\
 & Plain & OCtx & RCtx \\
 \cmidrule(lr){2-4}\\
SVL & .993 $\pm$ .015 & .997 $\pm$ .007 & .997 $\pm$ .006 \\
 & -- & -- & -- \\
SVR & .989 $\pm$ .014 & .992 $\pm$ .009 & .993 $\pm$ .009 \\
 & -- & -- & -- \\
NB & .926 $\pm$ .033 & .942 $\pm$ .029 & .940 $\pm$ .027 \\
 & -- & {\scriptsize 1} & {\scriptsize 1} \\
NN & .993 $\pm$ .012 & .990 $\pm$ .010 & .987 $\pm$ .013 \\
 & -- & -- & -- \\
RF & .971 $\pm$ .020 & .970 $\pm$ .021 & .976 $\pm$ .012 \\
 & -- & -- & -- \\
\end{tabular}%
\begin{tabular}{lll}
  \multicolumn{3}{c}{ZO} \\
  Plain & OCtx & RCtx \\
  \cmidrule(lr){1-3}\\
 .991 $\pm$ .021 & .995 $\pm$ .009 & .996 $\pm$ .009 \\
  -- & -- & -- \\
 .984 $\pm$ .022 & .990 $\pm$ .010 & .990 $\pm$ .013 \\
  -- & -- & -- \\
 .891 $\pm$ .049 & .924 $\pm$ .038 & .916 $\pm$ .040 \\
  -- & {\scriptsize 1} & {\scriptsize 1} \\
 .989 $\pm$ .018 & .987 $\pm$ .011 & .984 $\pm$ .015 \\
  -- & -- & -- \\
 .957 $\pm$ .029 & .962 $\pm$ .027 & .968 $\pm$ .018 \\
  -- & -- & -- \\
\end{tabular}
\end{table}

\begin{table}[htb]
\renewcommand{\arraystretch}{0.6}
\centering
\footnotesize
\caption{Classification quality for dataset 3, subject 4.\label{table:Set3_patient4_res}}
\begin{tabular}{llll}
 & \multicolumn{3}{c}{SqCov} \\
 & Plain & OCtx & RCtx \\
 \cmidrule(lr){2-4}\\
SVL & .964 $\pm$ .035 & .977 $\pm$ .018 & .971 $\pm$ .025 \\
 & -- & -- & -- \\
SVR & .972 $\pm$ .029 & .980 $\pm$ .017 & .975 $\pm$ .022 \\
 & -- & -- & -- \\
NB & .829 $\pm$ .055 & .888 $\pm$ .047 & .866 $\pm$ .041 \\
 & -- & {\scriptsize 1,3} & {\scriptsize 1} \\
NN & .895 $\pm$ .043 & .946 $\pm$ .020 & .926 $\pm$ .030 \\
 & -- & {\scriptsize 1} & -- \\
RF & .898 $\pm$ .052 & .915 $\pm$ .040 & .914 $\pm$ .044 \\
 & -- & -- & -- \\
\end{tabular}%
\begin{tabular}{lll}
  \multicolumn{3}{c}{ZO} \\
  Plain & OCtx & RCtx \\
  \cmidrule(lr){1-3}\\
 .947 $\pm$ .050 & .970 $\pm$ .020 & .964 $\pm$ .031 \\
  -- & -- & -- \\
 .957 $\pm$ .043 & .977 $\pm$ .019 & .967 $\pm$ .031 \\
  -- & -- & -- \\
 .759 $\pm$ .074 & .862 $\pm$ .067 & .827 $\pm$ .058 \\
  -- & {\scriptsize 1,3} & {\scriptsize 1} \\
 .850 $\pm$ .059 & .930 $\pm$ .024 & .906 $\pm$ .036 \\
  -- & {\scriptsize 1} & {\scriptsize 1} \\
 .855 $\pm$ .070 & .893 $\pm$ .051 & .891 $\pm$ .052 \\
  -- & -- & -- \\
\end{tabular}
\end{table}

\begin{table}[htb]
\renewcommand{\arraystretch}{0.6}
\centering
\footnotesize
\caption{Classification quality for dataset 3, subject 5.\label{table:Set3_patient5_res}}
\begin{tabular}{llll}
 & \multicolumn{3}{c}{SqCov} \\
 & Plain & OCtx & RCtx \\
 \cmidrule(lr){2-4}\\
SVL & .988 $\pm$ .016 & .991 $\pm$ .010 & .992 $\pm$ .010 \\
 & -- & -- & -- \\
SVR & .988 $\pm$ .012 & .992 $\pm$ .006 & .994 $\pm$ .007 \\
 & -- & -- & -- \\
NB & .817 $\pm$ .042 & .866 $\pm$ .040 & .845 $\pm$ .039 \\
 & -- & {\scriptsize 1,3} & {\scriptsize 1} \\
NN & .976 $\pm$ .023 & .980 $\pm$ .013 & .978 $\pm$ .013 \\
 & -- & -- & -- \\
RF & .945 $\pm$ .021 & .946 $\pm$ .016 & .943 $\pm$ .017 \\
 & -- & -- & -- \\
\end{tabular}%
\begin{tabular}{lll}
  \multicolumn{3}{c}{ZO} \\
  Plain & OCtx & RCtx \\
  \cmidrule(lr){1-3}\\
 .982 $\pm$ .025 & .988 $\pm$ .014 & .990 $\pm$ .013 \\
  -- & -- & -- \\
 .982 $\pm$ .018 & .990 $\pm$ .007 & .991 $\pm$ .009 \\
  -- & -- & -- \\
 .747 $\pm$ .057 & .825 $\pm$ .052 & .791 $\pm$ .057 \\
  -- & {\scriptsize 1,3} & {\scriptsize 1} \\
 .964 $\pm$ .034 & .974 $\pm$ .018 & .968 $\pm$ .017 \\
  -- & -- & -- \\
 .919 $\pm$ .030 & .926 $\pm$ .022 & .920 $\pm$ .022 \\
  -- & -- & -- \\
\end{tabular}
\end{table}

\begin{table}[htb]
\renewcommand{\arraystretch}{0.6}
\centering
\footnotesize
\caption{Classification quality for dataset 3, subject 6.\label{table:Set3_patient6_res}}
\begin{tabular}{llll}
 & \multicolumn{3}{c}{SqCov} \\
 & Plain & OCtx & RCtx \\
 \cmidrule(lr){2-4}\\
SVL & 1.00 $\pm$ .000 & 1.00 $\pm$ .000 & 1.00 $\pm$ .000 \\
 & -- & -- & -- \\
SVR & .996 $\pm$ .008 & .998 $\pm$ .003 & .997 $\pm$ .006 \\
 & -- & -- & -- \\
NB & .991 $\pm$ .013 & .993 $\pm$ .010 & .993 $\pm$ .010 \\
 & -- & -- & -- \\
NN & .990 $\pm$ .014 & .994 $\pm$ .009 & .991 $\pm$ .010 \\
 & -- & -- & -- \\
RF & .984 $\pm$ .022 & .984 $\pm$ .019 & .981 $\pm$ .022 \\
 & -- & -- & -- \\
\end{tabular}%
\begin{tabular}{lll}
  \multicolumn{3}{c}{ZO} \\
  Plain & OCtx & RCtx \\
  \cmidrule(lr){1-3}\\
 1.00 $\pm$ .000 & 1.00 $\pm$ .000 & 1.00 $\pm$ .000 \\
  -- & -- & -- \\
 .994 $\pm$ .012 & .998 $\pm$ .004 & .997 $\pm$ .006 \\
  -- & -- & -- \\
 .986 $\pm$ .019 & .992 $\pm$ .013 & .990 $\pm$ .013 \\
  -- & -- & -- \\
 .985 $\pm$ .020 & .992 $\pm$ .013 & .991 $\pm$ .011 \\
  -- & -- & -- \\
 .977 $\pm$ .032 & .977 $\pm$ .026 & .970 $\pm$ .029 \\
  -- & -- & -- \\
\end{tabular}
\end{table}

\begin{table}[htb]
\renewcommand{\arraystretch}{0.6}
\centering
\footnotesize
\caption{Classification quality for dataset 3, subject 7.\label{table:Set3_patient7_res}}
\begin{tabular}{llll}
 & \multicolumn{3}{c}{SqCov} \\
 & Plain & OCtx & RCtx \\
 \cmidrule(lr){2-4}\\
SVL & .991 $\pm$ .013 & .991 $\pm$ .014 & .988 $\pm$ .011 \\
 & -- & -- & -- \\
SVR & .977 $\pm$ .016 & .986 $\pm$ .010 & .984 $\pm$ .013 \\
 & -- & -- & -- \\
NB & .923 $\pm$ .022 & .942 $\pm$ .021 & .938 $\pm$ .022 \\
 & -- & {\scriptsize 1} & -- \\
NN & .943 $\pm$ .028 & .960 $\pm$ .015 & .954 $\pm$ .022 \\
 & -- & {\scriptsize 1} & -- \\
RF & .922 $\pm$ .026 & .940 $\pm$ .020 & .936 $\pm$ .026 \\
 & -- & -- & -- \\
\end{tabular}%
\begin{tabular}{lll}
 \multicolumn{3}{c}{ZO} \\
 Plain & OCtx & RCtx \\
 \cmidrule(lr){1-3}\\
 .986 $\pm$ .019 & .987 $\pm$ .016 & .982 $\pm$ .015 \\
-- & -- & -- \\
 .966 $\pm$ .024 & .981 $\pm$ .013 & .979 $\pm$ .016 \\
  -- & -- & -- \\
 .885 $\pm$ .033 & .921 $\pm$ .030 & .918 $\pm$ .030 \\
  -- & {\scriptsize 1} & {\scriptsize 1} \\
 .916 $\pm$ .040 & .944 $\pm$ .021 & .940 $\pm$ .027 \\
  -- & {\scriptsize 1} & {\scriptsize 1} \\
 .886 $\pm$ .038 & .917 $\pm$ .028 & .922 $\pm$ .033 \\
  -- & {\scriptsize 1} & {\scriptsize 1} \\
\end{tabular}
\end{table}

\begin{table}[htb]
\renewcommand{\arraystretch}{0.6}
\centering
\footnotesize
\caption{Classification quality for dataset 3, subject 8.\label{table:Set3_patient8_res}}
\begin{tabular}{llll}
 & \multicolumn{3}{c}{SqCov} \\
 & Plain & OCtx & RCtx \\
 \cmidrule(lr){2-4}\\
SVL & 1.00 $\pm$ .000 & 1.00 $\pm$ .001 & .999 $\pm$ .004 \\
 & -- & -- & -- \\
SVR & .998 $\pm$ .004 & .997 $\pm$ .004 & .998 $\pm$ .004 \\
 & -- & -- & -- \\
NB & .930 $\pm$ .021 & .940 $\pm$ .021 & .937 $\pm$ .021 \\
 & -- & {\scriptsize 1} & -- \\
NN & .987 $\pm$ .011 & .991 $\pm$ .007 & .994 $\pm$ .005 \\
 & -- & -- & -- \\
RF & .968 $\pm$ .016 & .978 $\pm$ .010 & .974 $\pm$ .017 \\
 & -- & -- & -- \\
\end{tabular}%
\begin{tabular}{lll}
  \multicolumn{3}{c}{ZO} \\
  Plain & OCtx & RCtx \\
  \cmidrule(lr){1-3}\\
1.00 $\pm$ .000 & 1.00 $\pm$ .001 & .998 $\pm$ .007 \\
  -- & -- & -- \\
 .997 $\pm$ .006 & .996 $\pm$ .006 & .997 $\pm$ .006 \\
  -- & -- & -- \\
 .897 $\pm$ .029 & .919 $\pm$ .030 & .915 $\pm$ .030 \\
  -- & {\scriptsize 1} & {\scriptsize 1} \\
 .982 $\pm$ .016 & .988 $\pm$ .010 & .991 $\pm$ .006 \\
  -- & -- & -- \\
 .953 $\pm$ .023 & .970 $\pm$ .012 & .965 $\pm$ .023 \\
  -- & -- & -- \\
\end{tabular}
\end{table}

\begin{table}[htb]
\renewcommand{\arraystretch}{0.6}
\centering
\footnotesize
\caption{Classification quality for dataset 3, subject 9.\label{table:Set3_patient9_res}}
\begin{tabular}{llll}
 & \multicolumn{3}{c}{SqCov} \\
 & Plain & OCtx & RCtx \\
 \cmidrule(lr){2-4}\\
SVL & .801 $\pm$ .033 & .836 $\pm$ .039 & .826 $\pm$ .043 \\
 & -- & {\scriptsize 1} & {\scriptsize 1} \\
SVR & .803 $\pm$ .057 & .850 $\pm$ .050 & .832 $\pm$ .062 \\
 & -- & {\scriptsize 1,3} & {\scriptsize 1} \\
NB & .528 $\pm$ .039 & .602 $\pm$ .041 & .581 $\pm$ .042 \\
 & -- & {\scriptsize 1,3} & {\scriptsize 1} \\
NN & .663 $\pm$ .045 & .735 $\pm$ .040 & .710 $\pm$ .037 \\
 & -- & {\scriptsize 1,3} & {\scriptsize 1} \\
RF & .675 $\pm$ .044 & .721 $\pm$ .057 & .724 $\pm$ .048 \\
 & -- & -- & {\scriptsize 1} \\
\end{tabular}%
\begin{tabular}{lll}
 \multicolumn{3}{c}{ZO} \\
  Plain & OCtx & RCtx \\
  \cmidrule(lr){1-3}\\
.715 $\pm$ .044 & .783 $\pm$ .049 & .766 $\pm$ .059 \\
  -- & {\scriptsize 1} & {\scriptsize 1} \\
 .719 $\pm$ .076 & .797 $\pm$ .065 & .769 $\pm$ .082 \\
  -- & {\scriptsize 1,3} & {\scriptsize 1} \\
 .378 $\pm$ .047 & .506 $\pm$ .049 & .475 $\pm$ .051 \\
  -- & {\scriptsize 1,3} & {\scriptsize 1} \\
 .535 $\pm$ .056 & .656 $\pm$ .053 & .621 $\pm$ .050 \\
  -- & {\scriptsize 1,3} & {\scriptsize 1} \\
 .553 $\pm$ .056 & .638 $\pm$ .063 & .640 $\pm$ .058 \\
  -- & {\scriptsize 1} & {\scriptsize 1} \\
\end{tabular}
\end{table}

\begin{table}[htb]
\renewcommand{\arraystretch}{0.6}
\centering
\footnotesize
\caption{Classification quality for signalset 3. Average ranks over all subjects\label{table:Set3_avg_ranks}}

\begin{tabular}{llll}
 & \multicolumn{3}{c}{SqCov} \\
 & Plain & OCtx & RCtx \\
\cmidrule(lr){2-4}\\

SVL & 1.667 & 2.333 & 2.000 \\
 & -- & -- & -- \\
SVR & 1.333 & 2.444 & 2.222 \\
& -- & {\scriptsize 1} & {\scriptsize 1} \\
NB & 1.000 & 3.000 & 2.000 \\
& -- & {\scriptsize 1,3} & {\scriptsize 1} \\
NN & 1.333 & 2.778 & 1.889 \\
 & -- & {\scriptsize 1,3} & {\scriptsize 1} \\
RF & 1.444 & 2.556 & 2.000 \\
 & -- & {\scriptsize 1} & {\scriptsize 1} \\
\end{tabular}%
\begin{tabular}{lll}
  \multicolumn{3}{c}{ZO} \\
 Plain & OCtx & RCtx \\
\cmidrule(lr){1-3}\\
 1.667 & 2.333 & 2.000 \\
  -- & -- & -- \\
 1.333 & 2.556 & 2.111 \\
  -- & {\scriptsize 1} & {\scriptsize 1} \\
 1.000 & 3.000 & 2.000 \\
  -- & {\scriptsize 1,3} & {\scriptsize 1} \\
 1.222 & 2.778 & 2.000 \\
  -- & {\scriptsize 1,3} & {\scriptsize 1} \\
 1.111 & 2.556 & 2.333 \\
  -- & {\scriptsize 1} & {\scriptsize 1} \\
\end{tabular}

\end{table}

\begin{table}[htb]
\renewcommand{\arraystretch}{0.6}
\centering
\footnotesize
\caption{Classification quality for signalset 3. Average ranks over all subjects and base classifiers\label{table:Set3_avg_ranks_c}}

\begin{tabular}{llll}
& \multicolumn{3}{c}{SqCov} \\
 & Plain & OCtx & RCtx \\
\cmidrule(lr){2-4}\\

Avg Rnk & 1.356 & 2.622 & 2.022 \\
& -- & {\scriptsize 1,3} & {\scriptsize 1} \\
\end{tabular}%
\begin{tabular}{lll}
  \multicolumn{3}{c}{ZO} \\
   Plain & OCtx & RCtx \\
\cmidrule(lr){1-3}\\
 1.267 & 2.644 & 2.089 \\
 -- & {\scriptsize 1,3} & {\scriptsize 1} \\
\end{tabular}

\end{table}

To summarise, the answers to the research questions are as follows. \textbf{RQ1} In terms of classification quality, context-dependent methods tend to be similar or better than the context-free model. However, their true strength is that they can preserve the classification quality of the context-free method and significantly improve the dexterity of the prosthesis.   \textbf{RQ2} Generally, the optimisation of the box structure allows us to achieve slightly better classification quality when compared to the random choice of the movement-to-class binding. Consequently, the optimisation process should be performed when building the classifier for the prosthesis. 

\FloatBarrier
 
\section{Discussion}\label{sec:discussion}
In this section, we will provide a discussion of the obtained experimental results against the state-of-the-art work presented in the introductory section. For two reasons, however, this will not be a literal quantitative comparative analysis of the developed method and related methods in the literature. First, as mentioned in the introductory section, the proposed context-dependent method does not have strict counterparts described in the literature. Consequently, the quality measures used in this work are problem-specific and cannot be directly compared to the results presented by other researchers. Secondly, the experimental datasets used in the papers cited in this section are not publicly available, which makes a direct comparison impossible. Although the discussion is necessarily descriptive, we will present conclusions, observations, and qualitative or indicative quantitative comparisons, allowing us to highlight the advantages, disadvantages, similarities, and differences of the analysed methods.

\subsection{The context-dependent recognition system as a multiclassifier system (MCS)}\label{sec:discussion:MCS}

The context-dependent recognition system proposed in this paper is a multiclassifier system in which the initial classifier and the box classifiers are the base classifiers. The proposed approach for context recognition classifies an incoming object using a single selected classifier. Thus, it adheres to the dynamic classifier selection procedure (DCS). A local, context-dependent recognition task is defined by the box-classifier that was selected. Table~\ref{table:MCSComparison} displays the related work that involves ensemble methods applied to sEMG-based recognition of upper/lower limb movements.

The meanings of the columns are as follows:
\begin{itemize}
    \item Ref: Citation of the reference paper. For rows presenting the system developed in this work, the signalset number is also given. 

    \item Method: A short description of the MCS method used.

    \item Objective: A short description of the objective of the study.

    \item Number of:
        \begin{itemize}
            \item Cl: The number of classes.
            \item Ch: The number of sEMG channels (n/a -- not available)
            \item Subj: The number of subjects. H stands for able-bodied subjects and A stands for amputees.
        \end{itemize}
    \item BC: Base classifiers used.

    \item Difference:
        \begin{itemize}
            \item Max/Min: Maximum / minimum accuracy difference between the committee and the best single classifier. This value must be multiplied by $10^{-2}$ to get the exact value. If the difference is negative, it means that the single classifier outperforms the MCS system. Values in parentheses indicate for which base classifier the difference is observed. For rows related with the method presented in this paper, instead of accuracy, the zero-one measure is considered.
        \end{itemize}
\end{itemize}

The table presents various multiclassifier systems including the well-known methods of Bagging and Boosting. The purpose of presenting these results is to demonstrate that the use of multiclassifier systems can improve the classification quality of the system compared to single-classifier systems. Various base classifiers were used to create multiclassifier systems in the presented methods. Different datasets were used to obtain the results, with different numbers of classes and EMG channels. Data were collected from able-bodied individuals and amputees. We present the lowest and highest difference in classification accuracy between a classifier committee and a single classifier as a measure of improvement. The classifier committee performs better than the single one when the difference is positive. If the difference is negative, a single classifier system performs better. The results presented show that a multiclassifier system is not always superior to a single one, but we can find conditions under which the ensemble may perform better for each study. For comparison, we also presented our results in the table. However, it must be emphasised that in the case of our result a different, more restrictive zero-one criterion is used. This criterion is similar to the accuracy score, but it counts the number of correctly classified movement sequences instead of individual gestures. Our results also confirm that, in general, using a classifier committee instead of a single classifier makes room for improvement in classification quality. It is important to emphasise that our primary objective with the multiclassifier system is to improve prosthesis dexterity by increasing the number of movements that can be embedded into a prosthesis. The overall improvement of classification quality is a side effect of employing a context-dependent classifier committee. However, this improvement is also important. The method presented in this paper follows the trends that may be observed for the reference methods. That is, the use of the ensemble system can increase the classification quality when compared to the single-classifier system.

\begin{table}[htb]
\centering
\footnotesize

\caption{Comparison of the proposed model with the related multiclassifier systems\label{table:MCSComparison}}
\begin{tabular}{p{1.5em}|p{10em}|p{10em}|p{1em}|p{1em}|p{2em}|p{8em}|p{3.5em}|p{3.5em}}
& & &\multicolumn{3}{c|}{Number of}& &\multicolumn{2}{c}{Difference}\\
Ref & Method & Objective & Cl & Ch & Subj & BC & Max ($10^{-2}$) & Min ($10^{-2}$) \\
\toprule

\cite{Kurzynski2017} & Heterogeneous DES scheme with original competence measure & sEMG-based hand movement recognition &6 &  8 & 2(H)& LDA, QDA, NM, kNN, NB, DT, MLP& 3.6 & 1.9 \\

\hline
\cite{Subasi2020}& Bagging & sEMG-based hand movement recognition& 6 & n/a & 5(H)& SVM, kNN, DT, NB, RF& 2.67 (DT) & -4.88 (kNN) \\

\hline
\cite{Subasi2020}& Boosting (Adaboost) & sEMG-based hand movement recognition& 6 & n/a & 5(H)& SVM, kNN, DT, NB, RF& 3.11 (SVM) & 0.11 (RF)\\

\hline
\cite{Akbulut2022}& Bagging & sEMG-based recognition of the phantom movements for the upper and lower limbs.  & 2 -- 4 & 2& 25(H), 38(A)& SVM, DT, kNN& 10.91 (DT) & -2.58 (kNN)\\

\hline
\cite{Akbulut2022}& Boosting (Adaboost) & sEMG-based recognition of the phantom movements for the upper and lower limbs.  & 2 -- 4 & 2& 25(H), 38(A)& SVM, DT, kNN& 4.33 (SVM) & -4.13 (SVM)\\

\hline
\cite{Freitas2023}& SCS scheme based on the single best approach & Recognition of surgical instrument signaling gestures  & 4 & 8& 10(H)& kNN, RF, SVM, MLP, LDA, QDA & 4.00 (SVM) & -3.00 (SVM)\\

\hline
This work, set 1& DCS scheme with original selection procedure &sEMG control of bionic upper limb prosthesis  & 8 & 8& 1(H)& SVL, SVR, NB, NN, RF & 17.70 (NN) & 1.10 (SVL)\\

\hline
This work, set 2& DCS scheme with original selection procedure &sEMG control of bionic upper limb prosthesis  & 5 & 8& 1(A)& SVL, SVR, NB, NN, RF & 10.70 (NN) & 0.00 (RF)\\

\hline
This work, set 3& DCS scheme with original selection procedure &sEMG control of bionic upper limb prosthesis  & 6 & 8--12& 9(A)& SVL, SVR, NB, NN, RF & 12.10 (NN) & -0.70 (RF) \\

\bottomrule

\end{tabular}

\end{table}

\subsection{Context-dependent recognition system versus targeted muscle reinnervation surgery as a tool to enhance the dexterity of the bionic prosthetic hand }\label{sec:discussion:Context}

The most important advantage of the proposed context recognition system is the increase in the repertoire of movements (grasping and manipulation) controlled by the user. Improved control of the prosthetic hand can also be achieved by targeted muscle reinnervation (TMR) surgery, where amputated nerves are transferred to reinnervate new muscle targets in the stump. The work~\cite{Simon2023} presents the results of clinical trials conducted on a group of 7 people who underwent TMR treatment. The effectiveness of treatment was assessed by comparing clinical outcome measures (the Southampton Hand Assessment Procedure, the Jebsen-Taylor Hand Function Test, the Assessment of Capacity for Myoelectric Control). The experimental protocol required that all subjects had used the commercial hand prosthesis both before and after TMR surgery during 8-week home trials. For all subjects, an improvement was observed. Because our experimental research did not involve clinical trials, the results obtained can only be linked to the coefficient called 'the number of configured grips'. The study reports that the average value of this measure increased from 3.7 before TMR surgery to 4.1 after treatment. Furthermore, an offline electromyography analysis showed a decrease in the grip classification error post-TMR surgery compared to pre-TMR surgery. For example, for the four grasp classification problem and the four-channel EMG data, the classification error was reduced from 3.2\% to 2.4\%. For the method proposed in this paper, the number of available grasps and manipulations is doubled. The increase may be even greater if the user accepts more than two interpretations of a single muscle-activation pattern. The proposed multiclassifier system can also improve the quality of classification. The use of TMR surgery is limited by several inclusion criteria that must be met before surgery, as it is a highly invasive method. Fifteen people were chosen for the research described in~\cite{Simon2023}, but only nine of them underwent TMR treatment due to restrictive inclusion criteria. The method described in this paper is non-invasive. As a result, patients are not subjected to restrictive inclusion criteria.

\subsection{Biomimetic and non-biomimetic (arbitrary) control startegies}\label{sec:discussion:biomimetic}

Biomimetic control is often used in electromyographic control systems for bionic prosthetic hands~\cite{Chen2023}. It means that the user's intention to move the prosthesis and the imagination of moving the phantom limb are the same. We abandon this assumption in the developed method to distinguish between the intention of moving the prosthesis and the imagination of moving the phantom hand. In other words, the imagination of the phantom hand movement does not have (but it may) be identical to the desired movement of the prosthesis. This control strategy is called non-biomimetic control or arbitrary control~\cite{Schone2023}. This arbitrary control strategy does not make sense when applied to the problem of controlling a robotic hand by able-bodied operators. For example, it would be troublesome to apply the non-biomimetic strategy in tele-surgery. In the case of an able-bodied operator, the biomimetic control strategy is more natural and intuitive. This is because the intended movement of the hand must be copied by the robotic hand~\cite{Rudiman2023}. In the case of people with amputation, the situation is different. There is no hand movement, there is only an imagination of the phantom hand moving, and only the user knows what the phantom hand is doing. So why wouldn't the user have the right to freely shape this imagination and freely associate it with the intention of moving the prosthesis, taking into account only their own comfort, intuition, and individual abilities of activating the residual limb muscles? Consequently, for the amputees, there is a great room for using non-biomimetic control strategy. The reason for this is the distinction between the intention of the prosthesis movement and the imagination of the phantom hand movement. This observation is the basis for the proposed context-dependent system. Indeed, in our system we propose to link single imagination of the phantom hand movement to multiple gestures of the prosthetic hand.

The paper~\cite{Schone2023} presents the results of the experimental study aimed at comparing biomimetic and arbitrary control strategies. The research involved 60 able-bodied people. Three separate groups were created: a control group, a mimetic training group, and an arbitrary training group. Subjects used i-Limb Quantum~\footnote{\url{https://www.ossur.com/en-us/prosthetics/arms/i-limb-quantum}} bionic hand, which was controlled by the Coapt pattern recognition controller~\footnote{\url{https://coaptengineering.com/technology}}. There are three main findings of the conducted research: (1) biomimetic and arbitrary control show similar increases in bionic hand embodiment; (2) biomimetic control provides some early training speed benefits but this advantage reduces with more training; (3) arbitrary strategy increases generalisation to new control mappings. Therefore, the practical thesis is justified that the context-dependent recognition system facilitates increasing the dexterity of the prosthesis by adding new movements to the repertoire and creating new ideas about the movement of the phantom hand. There is also a justified recommendation that in case of difficulties in mastering the proposed control strategy with context recognition, the user's training time should be extended. In the proposed method, it is also possible to optimise the box structure for the multiclassifier system. The optimisation process is described in detail in Section~\ref{sec:contextdeprecsystem:optimization}. The initial box-structure optimisation process allows the classifier ensemble to be most optimally adjusted to the repertoire of intentions to perform movements that the patient is capable of generating. The difference in training speed between biomimetic and non-biomimetic strategies can be reduced, at least partially.

\subsection{System for control of bionic hand prosthesis as a finite state machine }\label{sec:discussion:FSM}

In this work, we proposed a context-dependent classification method to recognise grasping intentions by analysing sEMG signals. The classification method is used in the task of controlling upper limb prostheses. The context is defined via a nested box structure. This structure describes the sequences of movements that allow the user to change the inner state of the prosthesis. An equivalent description of the proposed method may be formulated using a finite-state machine (FSM). A separate classifier is related to each prosthesis state. Each of the state-related classifiers may employ a different model-building strategy, a different set of attributes, and a different set of classes to predict. The state transitions are triggered when the class-related classifier predicts a class. Then the state is changed, and another classifier is used to predict the grasping intention. In other words, the user can generate a sequence of muscle contractions to change the state of the prosthesis.

Describing the behaviour of the prosthesis or any other sEMG-controlled system using FMA is not a new idea. A key difference between methods that employ FSMs is the way in which the state change is triggered. Some researchers use simple but robust thresholding methods to trigger the state change when the selected muscle is activated~\cite{Nacpil2019}. However, for such methods, the number of patterns that the operator can generate is limited. The other way is to use a kind of GUI application to allow the user to change the prosthesis state manually~\cite{Fajardo2021}. However, this forces the user's attention to be divided between generating sEMG patterns and using the application at the same time. To alleviate this disadvantage, a more user-friendly interface can be used. In~\cite{Shi2023} the authors proposed to use augmented reality goggles together with eye movement tracking. However, user attention is split between sEMG generation and state switching. To overcome this issue, some authors proposed using sensors different from sEMG sensors to switch the state. The authors of~\cite{Fajardo2018} propose placing a webcam on the prosthesis. The webcam takes pictures of the items to be grasped, and a separated classifier based on a convolutional neural network classifies the item and switches the state of the prosthesis. Although this approach significantly increases the convenience of using the prosthesis, it also increases the cost of the device itself and its demand for power. Some authors proposed the use of simpler sensors such as gyroscopes to determine the state of the prosthesis~\cite{Patel2017}. The authors of \cite{Batzianoulis2019} proposed to use a goniometer to track changes in angular velocity of the elbow. Depending on the angular velocity, a different sEMG classifier is applied.  Other biosignals such as MMG can also be used to trigger the state change. Geng et al.~\cite{Geng2012} proposed a system in which the state/posture of the prosthesis is identified using an MMG signal, whereas the grasping movements are predicted using sEMG. In this approach, two classifiers are needed: one for posture prediction and the other for grasping movement prediction.

Other researchers, like us, decided to rely solely on the sEMG sensors installed in the prosthesis. In this case, the state transition is triggered by the decision of a classifier or classifiers trained using sEMG signals. Cardona et al.~\cite{Cardona2020} proposed a system that uses only one classifier to change the state of the prosthesis. This system may be considered equivalent to the \textbf{Plain} method that our methods are tested against. To verify the quality of classification offered by the method, the authors calculated the confusion matrix and the overall accuracy of a single classifier used in the system. They achieved an accuracy of 84.4\%. The authors did not evaluate the classification quality for the sequences of movements. This result cannot be directly compared with the context-dependent classifiers presented in this work. This is because the benchmark datasets used are completely different. The authors did not publish their dataset, so we are unable to perform a comparison on the same dataset.  However, in Section~\ref{sec:resultsanddisc}, we have shown that the context-dependent methods proposed in this paper may outperform the \textbf{Plain} method. It is worth noting that in our experiment a zero-one criterion was used, which is more restrictive than accuracy. The authors of \cite{DAccolti2023}  used a single SVM classifier to switch between states. Their approach involves continuous control of a chosen degree of freedom of the prosthesis in each state. They reported 82.1\% accuracy for able-bodied objects and 65.9\% accuracy for amputees. Authors of \cite{Piazza2020} also use a single classifier to change the state of the prosthesis, but the definition of the state is completely different. They define a two-dimensional state space based on the degree of actuation of the prosthesis. This approach allows the prosthesis to perform smooth transitions between predefined gestures. The conducted study was focused mainly on user experience. Consequently, the authors do not report the performance of the classifier, but rather measures related to user experience, such as the average time needed by a user to perform a certain task using the prosthesis.

\subsection{Context-dependent classification system versus single-classifier system}\label{sec:discussion:Single}

On the other hand, we are still able to compare our methods with single classifier systems tested on the dataset published by Al-Timemy et al.~\cite{AlTimemy2016}. For Random Forest, Naive Bayes and kNN classifiers and low force level, they obtained classification accuracy above 90\% (averaged over all test subjects). For the context-dependent methods presented in this paper, the average value of zero-one criterion is also above 90\%. As it was previously said, the zero-one criterion is much stricter than accuracy. This is because the entire object sequence must be classified correctly to count the sequence as correctly classified.

\section{Conclusions}\label{sec:conclusions}

In this paper, a novel method, which uses a context-dependent decision scheme is proposed for the recognition-based control of the upper limb bioprosthesis. The use of a properly shaped context allows for a different, but always unambiguous interpretation of the classification results, which increases the range of movements of the bioprosthesis under the user's control. Although this solution is less intuitive than the classical scheme, in which each class of sEMG signals always means one movement of the prosthesis, it has very practical properties and high application potential. Because intuitiveness is a matter of developing the user's \textit{imaginarium}, the appropriate training method may bring the moment of its practical use closer.

The proposed context-dependent recognition scheme leads to a multiclassifier system. This fact alone creates a space to improve the quality of classification. 
The structure (consisting of boxes) of the proposed context-dependent system allows for the additional optimisation of each of the base classifiers, which is yet another source of improving the quality of classification. We are dealing here with a discrete optimisation problem with constraints, for which the cardinality of the set of feasible solutions depends on the number of classes and the maximum order of boxes (box classifiers) in the context structure. For the considered examples, in experimental studies with 5 and 6 classes, this cardinality did not exceed several dozen, and it was therefore possible to base the problem-solving procedure on the exhaustive search scheme. For larger sets of admissible solutions (e.g. for the example with 8 classes), an optimisation procedure based on an evolutionary algorithm with permutation coding of individuals can be used.

It should be emphasised that the developed method of context-dependent recognition is not a general purpose method. It has been tailored to the problem of controlling the bioprosthesis of the hand. It uses the specificity of this problem and is not suitable for other applications. Therefore, it is a problem-oriented method, which the authors consider to be its important advantage.

In recent years, solutions to the hand bioprosthesis control system based on the invasive approach have become more and more popular. Such concepts can expand the options for prosthetic integration, myoelectric signal detection, and sensation restoration, but will not replace methods and algorithms for effective biosignal recognition. 

\bibliographystyle{model1-num-names} 
%
%
\bibliography{cas-refs}

\end{document}